\documentclass[journal]{IEEEtran}
\usepackage{cite}

\ifCLASSINFOpdf
\else
\fi

\usepackage{subcaption}
\usepackage{graphicx}
\usepackage{amsfonts}
\usepackage{amsmath}
\usepackage{xfrac}
\usepackage{multirow}
\usepackage{booktabs}
\usepackage{balance}
\usepackage{url}

\usepackage[]{todonotes}

\begin{document}
%
\title{Multi-Agent Belief Sharing through \\Autonomous Hierarchical Multi-Level Clustering}


\author{Mirco~Theile,~\IEEEmembership{Student Member,~IEEE,}
        Jonathan~Ponniah,~\IEEEmembership{Member,~IEEE,}
        Or~Dantsker,~\IEEEmembership{Student Member,~IEEE,}
        and~Marco~Caccamo,~\IEEEmembership{Fellow,~IEEE}
\thanks{Mirco Theile (corresponding author), Or Dantsker, and Marco Caccamo are with Technical University of Munich (TUM), Garching, Germany, \{mirco.theile, or.dantsker, mcaccamo\}@tum.de}
\thanks{Jonathan Ponniah is with San Jose State University, San Jose, CA 95192, jonathan.ponniah@sjsu.edu}
\thanks{M.~Caccamo was supported by an Alexander von Humboldt Professorship endowed by the German Federal Ministry of Education and Research.}
}

%

\markboth{Submitted to IEEE Transactions on Robotics}%
{Theile \MakeLowercase{\textit{et al.}}: Multi-Agent Belief Sharing through Autonomous Hierarchical Multi-Level Clustering}


\maketitle

\begin{abstract}
Coordination in multi-agent systems is challenging for agile robots such as unmanned aerial vehicles (UAVs), where relative agent positions frequently change due to unconstrained movement. The problem is exacerbated through the individual take-off and landing of agents for battery recharging leading to a varying number of active agents throughout the whole mission. This work proposes autonomous hierarchical multi-level clustering (MLC), which forms a clustering hierarchy utilizing decentralized methods. Through periodic cluster maintenance executed by cluster-heads, stable multi-level clustering is achieved. The resulting hierarchy is used as a backbone to solve the communication problem for locally-interactive applications such as UAV tracking problems. Using observation aggregation, compression, and dissemination, agents share local observations throughout the hierarchy, giving every agent a total system belief with spatially dependent resolution and freshness. Extensive simulations show that MLC yields a stable cluster hierarchy under different motion patterns and that the proposed belief sharing is highly applicable in wildfire front monitoring scenarios. 
\end{abstract}


%
\IEEEpeerreviewmaketitle

\section{Introduction}

\IEEEPARstart{A}{utonomous} multi-agent systems are spurring advances in fields ranging from logistics to inspection, monitoring, surveillance, and forensics \cite{Guerrero2013,Bouafif2018,Kaleem2018,Ding2018}. Emerging applications of this technology include wildfire surveillance \cite{Julian,Haksar2018, viseras2021wildfire}, planetary exploration \cite{HASSANALIAN2018}, environmental surveying \cite{Nuske2015}, perishable data acquisition \cite{Meyer2015}, on-demand wireless provisioning \cite{Katila2017}, and aerial package delivery \cite{GOODCHILD2018}. These applications are specific instances of a more general multi-agent tracking problem. Agents on a two-dimensional grid must visit or track spatially distributed interest points that evolve over time. The interest points could signify burning trees, geological features, pick-up and delivery locations, or wireless customers. The system's objective is to discover or visit as many interest points as possible within some bounded time horizon. Since the costs and capabilities of commercial hardware show encouraging trends, the bottlenecks of this technology are increasingly software-related or algorithmic. This paper focuses on one such bottleneck; the coordination among multi-agent systems, addressed by incepting a backbone for communication and control policies. It improves, fully implements, and utilizes the concept presented in \cite{ponniah2021autonomous}.

\begin{figure}
    \centering
    \includegraphics[width=\columnwidth]{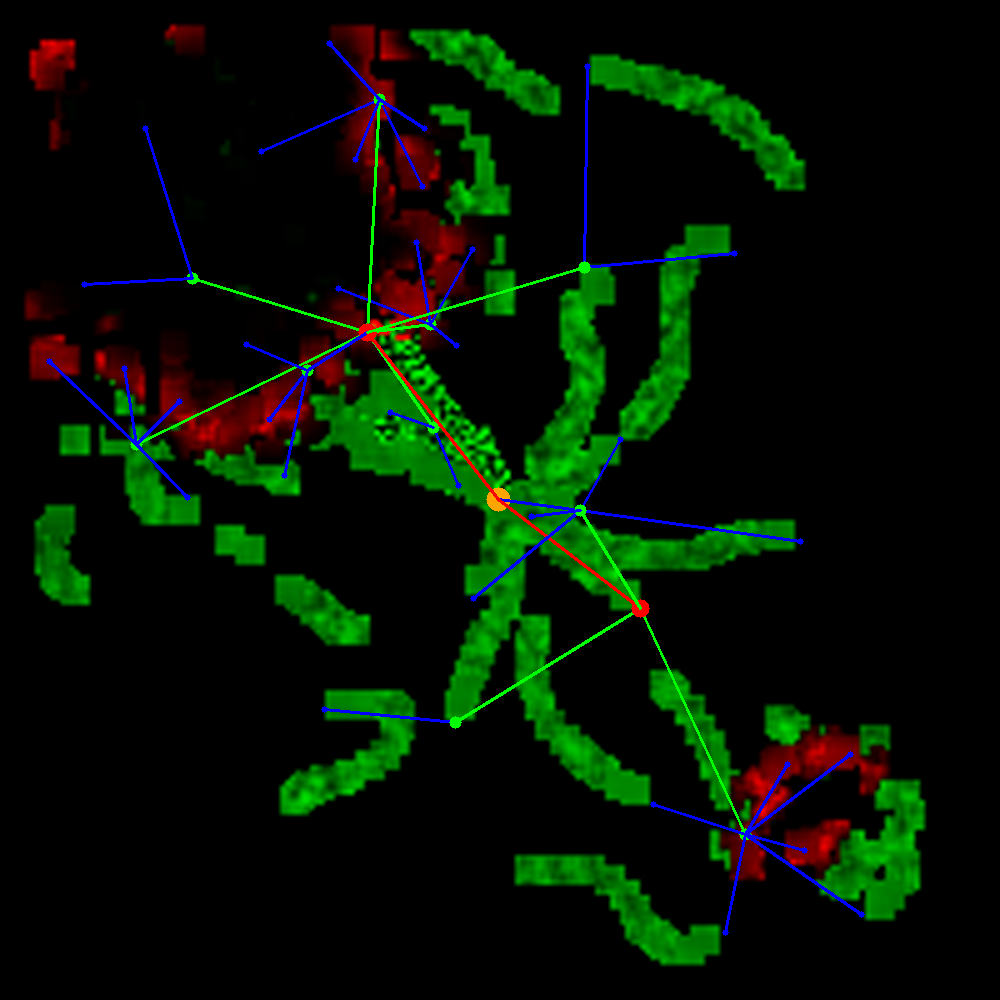}
    \caption{Example clustering of 50 agents during a wildfire monitoring mission, showing a base station's belief of two fires that is not older than 60 seconds. The different colors of the agents indicate cluster-head status and lines indicate cluster membership. The different resolutions of the fire (red) and trees (green) are resulting from data compression used in belief sharing.}
    \label{fig:clustering_example}
\end{figure}

Wildfire surveillance, the application considered in this paper, is an example of the multi-agent tracking problem that has received much recent attention.  Agents in this context are UAVs and interest points are burning trees. Each agent has direct access to its local observations, the measurements from onboard sensors. These measurements include the position and velocity of the agent as well as the images from its camera. The global state includes the positions and velocities of all agents and the position and status of each tree patch in the forest. The tree status indicates whether or not a tree is burning and the value of its remaining health. Through its camera, each agent can observe the status of the trees within a smaller sub-grid. The system's objective is to locate all the burning trees in the forest.  

To coordinate their activity, agents must first disseminate their observations, then use their shared observations to estimate the global state without relying on any pre-existing wireless infrastructure that could be absent in remote areas. This \textit{communication problem} is non-trivial if the number of agents is large or the agents are widely dispersed. The agents themselves must discover both the neighbors with which they can relay information and the topology of this induced network and collectively decide on the communication routes between all source-destination pairs. These routes must be updated constantly since the network topology changes with the agents' relative positions and motions.

In every time step, the agents must decide on a joint action given their shared view of the global system state such that the action sequence drives the system towards its objectives over the operating horizon. This \textit{control problem} is also non-trivial because its complexity scales poorly with the number of agents.  

We propose an autonomous hierarchical multi-agent multi-level clustering (MLC) strategy that addresses the communication problem and prepares the control problem for future work. For that, MLC induces a communication graph over which agents share their system belief, with the ultimate goal of achieving scalable system-wide coordination. The topology of the \textit{ideal} communication graph, which depends on the clustering metric, is time-varying since the agents are highly mobile. We propose a distributed protocol that adapts to the agents' motion, ensuring that the system remains connected through an appropriate clustering hierarchy, even if the agent trajectories are independent.

We utilize the induced graph for a novel belief sharing technique, which is based on the hypothesis that multi-agent tracking problems are locally-interactive; individual agents require precise observations only of the agents in the immediate vicinity but aggregated, compressed, and time-sampled observations of agents farther away. This hypothesis also applies to clusters of agents and clusters of clusters. The resolution and age of aggregate information that clusters receive from other clusters need only be precise and fresh for nearby clusters but can be compressed and older for remote ones. 

For that purpose, agents send their local observations to their cluster-heads. The cluster-heads aggregate the observations of their members, compress them, and send the aggregate belief to their cluster-heads. The process is repeated until the highest-level cluster-head is reached, from where the total aggregated and compressed belief is send down the hierarchy. Due to compression at every level, the belief of agents that are far away in the cluster hierarchy are more compressed than close ones. With MLC's induced topology, the quality of the shared belief is spatially dependent, making this technique ideally suited for coordination in locally-interactive applications. An example snapshot of this hierarchy on top of a belief of an active wildfire is depicted in Figure \ref{fig:clustering_example}.

\subsection{Contributions}

This work proposes a novel autonomous hierarchical multi-level clustering (MLC) technique to function as a backbone for multi-agent communication and coordination tasks. The main benefits of the proposed MLC are that it is scalable, does not require complicated consensus protocols, is applicable for homogeneous multi-agent systems, does not require a central entity with perfect knowledge of all agent positions, and does not require any motion directives. We further describe a methodology that uses the MLC backbone to aggregate local beliefs of agents and disseminate them to the other agents with spatially dependent compression. Agents can use the shared belief to make motion decisions, which, in this work, are utilized for monitoring stochastic wildfires. We show that the clustering significantly reduces the number of links and their distances. The spatially dependent compression shows to reduce data rate with little to no overall mission performance reduction. Furthermore, we describe a simple potential field method that utilizes the shared belief for wildfire front monitoring\footnote{A video showing the simulation and clustering explanation can be found here: \url{https://youtu.be/3s8DL-G6GAU}.}
, a good baseline comparison for future methodologies such as reinforcement learning. Simulations with potential field-induced motion primitives show that observation sharing through the MLC backbone achieves competitive \textit{fire-miss ratios} when compared with agents that perfectly share their observations with everyone. The main contributions can be summarized and include
\begin{itemize}
    \item a novel autonomous hierarchical multi-level clustering technique that functions as the backbone for multi-agent communication and coordination;
    \item a belief aggregation and dissemination method that uses the spatial nature of MLC for compression and its links for belief sharing;
    \item an analysis of different parameters and aspects of MLC through extensive simulations of different motion patterns, focusing on wildfire front monitoring.
\end{itemize}

The rest of this work is structured as follows: Section \ref{sec:clustering} describes MLC, giving a detailed description of cluster formation and maintenance. In Section \ref{sec:communication} the concept of meta beliefs is introduced, and methodologies for aggregation, compression, and dissemination are detailed. The evaluation methods are presented in \ref{sec:eval_methods}, followed by the evaluation results in \ref{sec:results}. Section \ref{sec:conclusion} concludes this work, giving a brief outlook to future work.

\section{Related Work}

The communication and the control problems have long but largely separate histories in the literature.  An important area of research in communication networks is multi-hop routing, finding paths over which nodes can relay messages between source-destination pairs.  The most widely used routing protocols discover the shortest or least-cost paths and fall under the \textit{link-state} \cite{McQuillan80} or  \textit{distance-vector} \cite{rfc1058} categories.  Link-state protocols such as  Open-Shortest-Path-First (OSPF) \cite{rfc2328} establish certificates between adjacent nodes (or agents) and disseminate (or flood) these certificates to all the agents in the system.  Each agent reconstructs the network topology from the link certificates and finds the shortest paths to all other agents using Dijkstra's algorithm \cite{Dijkstra1959}.  Distance-vector protocols such as Routing Information Protocol (RIP) \cite{rfc1058} avoid any flooding overhead; agents do not maintain a global view of the topology, but instead exchange distance-vectors with their neighbors and perform Bellman updates on the \textit{next-hop} and \textit{cost-to-go} destination entries in their forwarding tables.

Both link-state and distance-vector protocols have been adapted to the more nimble class of \textit{ad-hoc} wireless networks, which operate without fixed infrastructure and consist of power-limited wireless nodes.  The inherent instability of the topology makes link-state routing more cumbersome and leads to the count-to-infinity problem in distance-vector routing.  The former is addressed with Optimized Link State Routing (OLSR) \cite{Jacquet2001,rfc3626}, and the latter with Ad-Hoc On-Demand Distance-Vector (AODV) \cite{Perkins99} and Destination-Sequence-Distance-Vector (DSDV) routing \cite{Perkins1994}.

Routing protocols traditionally operate in the network layer (layer 3) independent of the application layer above (layer 7) and base their selection criteria on the shortest paths (as opposed to the most stable paths).  Multi-level clustering, by contrast, is a cross-layer design; the information profile best suited for the multi-agent tracking problem (in the application layer) determines the network topology and the communication routes.  Like distance-vector routing, agents maintain a limited view of the network topology, in this case, their individual clusters' status, to keep the overhead scalable.  However, protocol overhead is not the sole factor of concern in scalable multi-agent tracking; there are fundamental limits on the throughput that wireless networks can support, regardless of the protocol overhead \cite{Kumar2000}.  In particular, the normalized achievable source-destination data rate is sub-linear or worse in the number of nodes for all fading regimes \cite{Xie2004,Ozgur2007,Ghaderi2009}, thus implying that scalable multi-agent tracking is impossible without data compression.  These limits further justify a cross-layer design that combines data compression and routing.

Protocols can also be classified as either pro-active \cite{Jacquet2001,Perkins1994}  or reactive \cite{Johnson96dynamicsource,Perkins99}.  Pro-active protocols periodically update their routing tables to ensure the entries are always fresh. In contrast, reactive protocols only initiate updates when routes not offered in the existing tables are requested.  The protocol proposed in this paper pro-actively assimilates and transfers nodes within the clustering hierarchy using periodically scheduled maintenance phases.

Clustering is most commonly associated with wireless sensor networks and Vehicular Ad-Hoc Networks (VANETs).  The process of cluster formation is non-trivial and requires distributed consensus algorithms for agents to elect a cluster-head and agree on the cluster composition.  Variations of multi-level clustering have been proposed in \textit{endpoint-centric} mobile ad-hoc networks to reduce routing complexity.  We propose MLC for a fundamentally different reason; to facilitate coordination and information processing in \textit{data-centric} \textit{mobile} ad-hoc networks. This kind of strategy has been considered in data-centric ad-hoc networks with \textit{stationary} nodes.  The idea of \textit{in-network} data aggregation and processing generally appears in \textit{future network} architectures.  As far as the authors are aware, MLC has never been proposed for enabling coordination in multi-agent systems. 

Multi-agent tracking can be modeled as a Markov Decision Process (MDP) in which the optimal control policy maximizes the number of visited interest-points (i.e., the system return) \cite{Bellman}.  An extension of this framework called Partially Observable MDPs (POMDPs) introduces the notion of agent \textit{beliefs}; statistical observations correlated to the global state \cite{sondik,Kaelbling95planningand}.  Beliefs are useful when the global state is not itself directly observable.  MLC employs the notion of a \textit{meta-belief} which includes such metadata as the resolution of a compressed camera image and the time at which the image was taken.  Meta-data, a distinguishing feature of data-centric networking, helps synthesize a global view of the state from the compressed local views disseminated by the agents.   

Although optimal control policies are outside the scope of this paper, wildfire simulations based on potential-field based motion-primitives \cite{khatib1985} show that MLC achieves competitive fire-miss ratios, evidence of locally-interactive structure in the state-space \cite{Boutilier1996}.  This structure, which enables state and task decomposition, simplifies the planning and control problems \cite{STONE1999,simmons2007,Oliehoek2013,Amato}.  Recent work has also shown that data aggregation/compression can simultaneously reduce the complexity of multi-agent reinforcement learning and enable coordination.  We will build on these results by integrating reinforcement learning with MLC to solve for optimal control policies.  Multi-agent reinforcement learning with limited nearest-neighbor coordination has also been proposed for wildfire surveillance \cite{Haksar2018,Julian}.  

\section{Multi-Level Clustering}
\label{sec:clustering}
The following section describes fundamental clustering rules and concepts, which ensure that disconnected groups of decentralized, asynchronous agents can form a stable functioning network despite the differences in their take-off and landing times and their motion dynamics.  The proposed \textit{hierarchical multi-level clustering} approach takes agents from this initial uncoordinated state to a steady-state where full system-wide coordination is achievable. For now, the technique assumes that agents are in \textit{one-hop range} to each-other to guarantee that all agents are clustered. However, the results show empirically that this assumption can be relaxed. In the following, agents are denoted through $\alpha$.

\subsection{Decentralized to Hierarchical}
At the moment of spawning, which occurs at take-off, each agent knows of no other neighbors and belongs to no cluster. Hence, the immediate tasks are to discover whether any neighboring agents or agent clusters exist in the vicinity and to alert these agents of their presence. Newly spawned agents accomplish both tasks by first broadcasting "hello" messages that include their position while listening for other agents' "hello" messages, sharing this information. From the "hello" messages, agents create lists of their neighboring agents and clusters.

The next task is to either initiate cluster formation with neighboring unclustered agents or, if available, join a neighboring cluster that is not at capacity. After successfully joining or creating a cluster, coordination of communication is handled by a cluster-head. The cluster-head of this level-1 cluster is chosen according to its proximity to the centroid of the agents' positions in the cluster. Cluster-heads repeat the decentralized process, searching for other same-level cluster-heads to create a cluster of one level higher or joining existing clusters that are precisely one level higher. Repeated, this yields the desired hierarchy, in which cluster-heads can schedule internal communication, leading to more efficient synchronous communication schemes. As a definition, we define the highest cluster-head in a sub-tree as its boss. If the sub-tree encompasses all agents, then it is the entire tree, and its boss is the boss of all agents. Agents that are not cluster-heads are called idle.

\begin{figure}
    \centering
    \includegraphics[width=0.9\columnwidth]{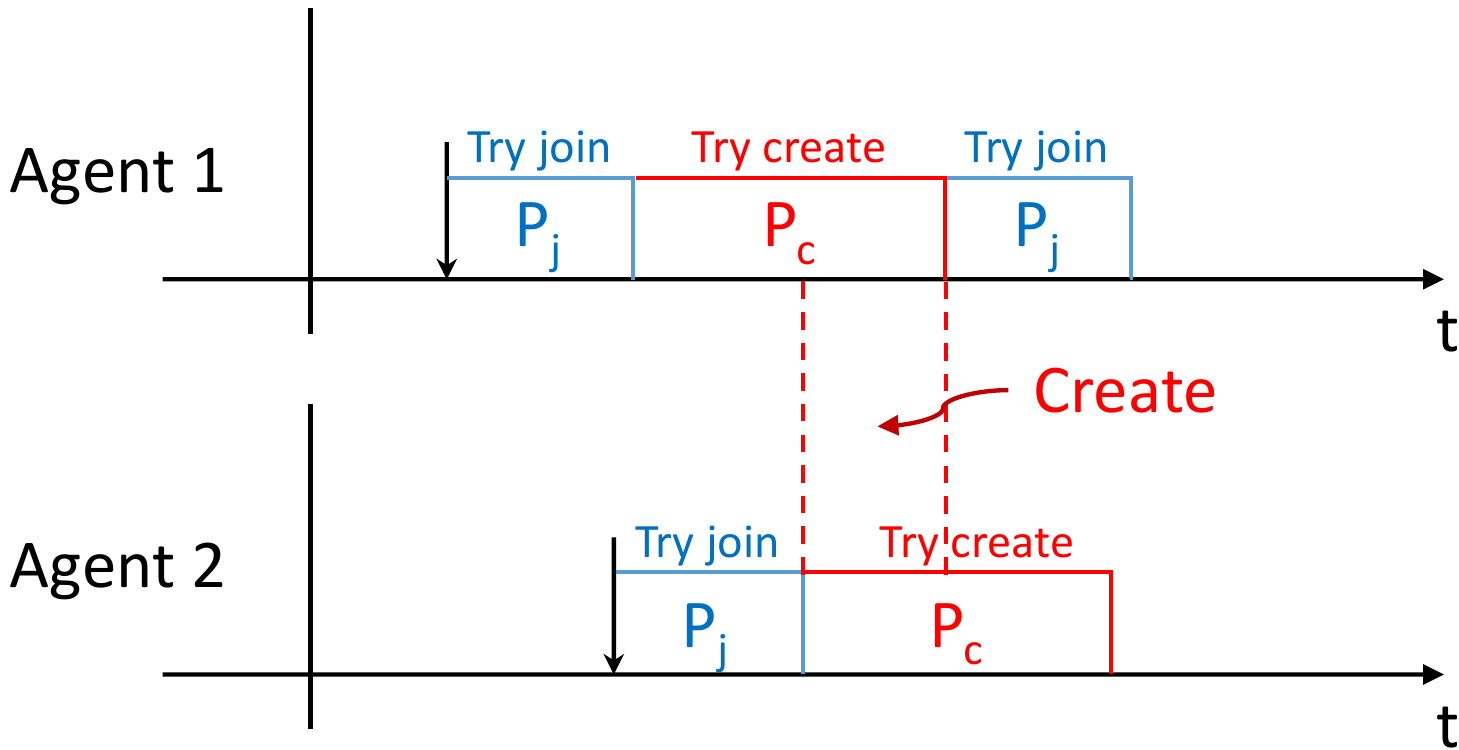}
    \caption{Scheduling of join and create task on individual agents, assuring overlap of create phases.}
    \label{fig:join_create_time}
\end{figure}
The decentralized clustering is achieved by alternating \textit{join} and \textit{create} tasks on agents as depicted in Figure \ref{fig:join_create_time}. In the \textit{join} task, agents try to join existing clusters, while in the \textit{create} task, they try to create a cluster with other agents that are also in their \textit{create} task. By increasing the length of the \textit{create} task, overlap can be guaranteed, allowing stable cluster formation from the beginning. Through spawning agents with the \textit{join} task, active overall clustering is improved since agents prefer joining existing clusters instead of creating new ones.

When creating a cluster, a \textit{cluster token} is created. The token describes the created cluster through its level, its members, and potentially higher-level cluster-head. Treating each cluster as a token that is owned only by the cluster-head allows for the concepts of token creation, passing, and destruction. With token creation, new clusters can be created. With token passing, the cluster-head can send the token information to another agent, making it the cluster-head of the cluster described by the token. With token destruction, clusters can be destroyed.

\subsection{Clustering Rules}
We define three rules for the proposed multi-level clustering to improve the applicability of the clustering in real-world environments. Their rationale and resulting consequences are elaborated in the following.

\subsubsection{Every agent can only be a cluster-head of one cluster.}
A cluster-head performs maintenance and cluster verification tasks to maintain and improve the clustering and tasks for aggregating and disseminating information to its members and superiors. These tasks require computational and communication capacities. If agents were allowed to be cluster-heads of multiple clusters, their onboard hardware would need to be provisioned for the worst-case computational and communication load. This worst-case does not have an upper bound if the number of agents and the operating area are unbounded because an agent could be a cluster-head of an unbounded number of clusters. Even if the number of agents and operating area are bounded, the worst-case load is still significantly higher than the average case load. In both cases, agents would be required to be drastically over-provisioned in computational hardware and communication resources. To avoid this, we do not allow agents to be cluster-heads of multiple clusters, balancing the loads more evenly.

\subsubsection{Every cluster-head has to be a member of a level-1 cluster.}
Every agent, cluster-head or idle, still conducts its regular mission tasks. Thus cluster-heads have to be members of level-1 clusters to receive and report the same level of aggregate information, allowing for homogeneous multi-agent systems, simplifying many planning and control problems. 

\subsubsection{A cluster-head of any level has to be a level-1 member of its sub-tree.}
The tree in this rule refers to all agents that are connected through multi-level clustering. The sub-tree of one cluster-head contains all of its descendants. This rule is an extension of \textit{Rule 2}, adding that the cluster-head has to be a leaf in its sub-tree. If agents are non-moving, this rule is enforced through \textit{Rule 1} and \textit{Rule 2}, as the agents can only be selected as cluster-heads if they are idle members of the tree. However, this rule is not automatically enforced if agents are moving, as they can change level-1 cluster membership, potentially leaving the sub-tree. Without \textit{Rule 3}, action directives decided by a cluster-head for its cluster may conflict with action directives it receives as a level-1 member of an external tree. This conflict creates ambiguity in action selection when applying the concept of joint actions for multi-agent planning. As joint actions will be an essential element of hierarchical planning, which is the ultimate goal of this research direction, \textit{Rule 3} eliminates the potential hurdle.

\subsection{Clustering Example}
\begin{figure}
    \centering
    \includegraphics[width=0.7\columnwidth]{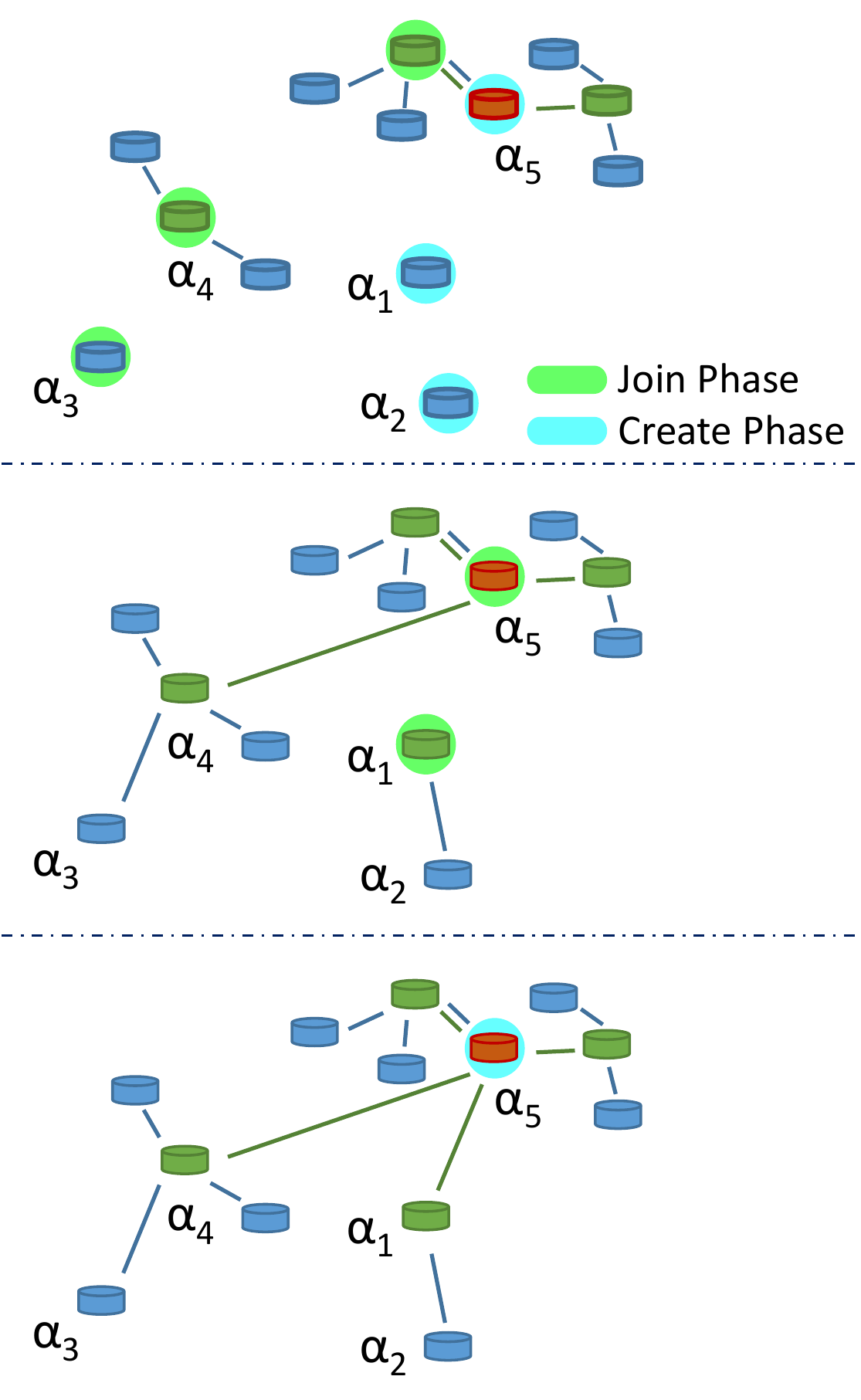}
    \caption{Three samples of the join-create based decentralized clustering.}
    \label{fig:join_create}
\end{figure}
\begin{figure}
    \centering
    \includegraphics[width=\columnwidth]{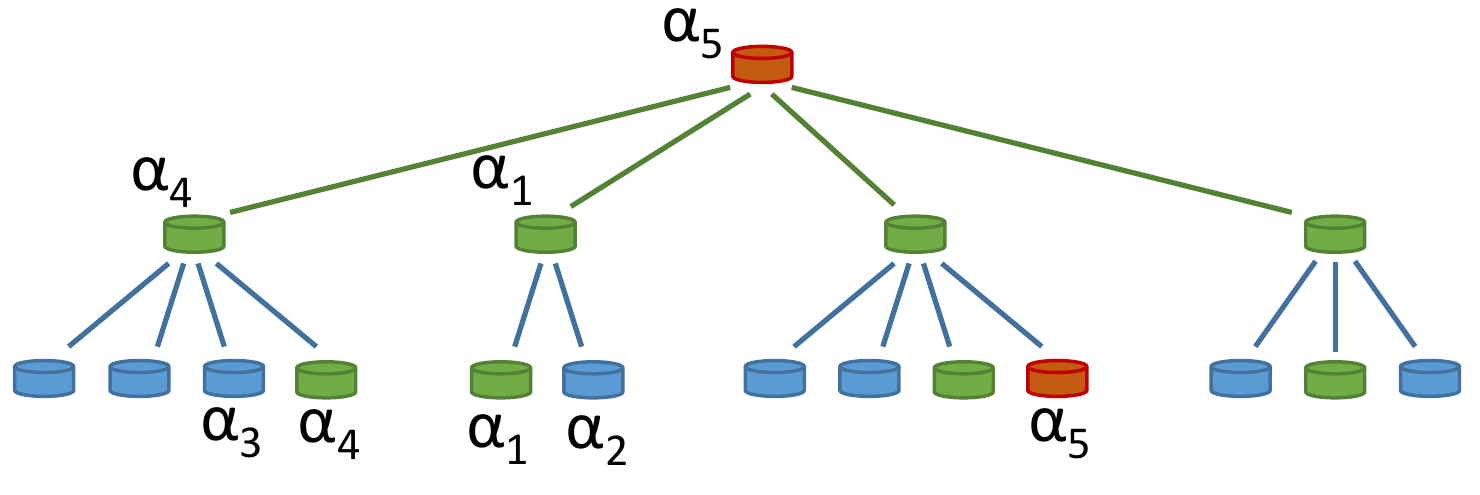}
    \caption{Tree view of the final clustering in Figure \ref{fig:join_create}}
    \label{fig:tree_view}
\end{figure}
An example of the decentralized clustering containing standard scenarios is depicted in Figure \ref{fig:join_create}. It shows three snapshots in time with different stages of the clustering. In the first part, $\alpha_1$-$\alpha_3$ are unclustered, $\alpha_4$ is a cluster-head of a level-1 cluster without a higher-level head, and $\alpha_5$ is a level-2 cluster-head without a higher-level cluster. All five agents are trying to form clusters decentralized and are thus either in their \textit{create} phase ($\alpha_1$,$\alpha_2$, and $\alpha_5$) or their \textit{join} phase ($\alpha_3$ and $\alpha_4$). Consequently, in the second snapshot, $\alpha_1$ and $\alpha_2$ formed a level-1 cluster, electing $\alpha_1$ as cluster-head. Note that the selection of the cluster-head in a cluster with two members is random. Since $\alpha_3$ was in the \textit{join} phase and $\alpha_4$ is close and not full, $\alpha_3$ joined $\alpha_4$'s level-1 cluster. Similarly, $\alpha_4$ joined $\alpha_5$'s level-2 cluster. Note that the cluster-head of the joined cluster does not need to be in its \textit{join} phase. The \textit{join} and \textit{create} phases are only relevant to agents of the same level. As there are no other level-2 cluster-heads, $\alpha_5$ toggles its decentralized phase to \textit{join}, trying to find level-3 cluster-heads to join. In the last snapshot, $\alpha_2$ joined $\alpha_5$'s level-2 cluster. $\alpha_5$ did not find any level-3 cluster to join and thus toggled back to its \textit{create} phase.

Figure \ref{fig:tree_view} shows the tree view of the last snapshot in Figure \ref{fig:join_create}. All three clustering rules can be observed in this. $\alpha_5$, the boss, is only a level-2 cluster-head, not also a level-1 head. All cluster-heads are level-1 members and specifically level-1 members in their sub-tree. 

\subsection{Cluster Maintenance}

After a cluster is formed, strategies for cluster maintenance can be handled by the cluster-heads in a locally centralized manner. We developed five maintenance strategies to improve and stabilize the clustering under relative motion, spawning, and despawning of agents. The following describes how cluster-heads can change, how agents can be transferred between clusters, how large clusters can split into two, how neighboring clusters can assimilate small clusters, and how the depth of the cluster tree can be regulated.

\subsubsection{Cluster-head Reassignment}
\begin{figure}
    \centering
    \includegraphics[width=\columnwidth]{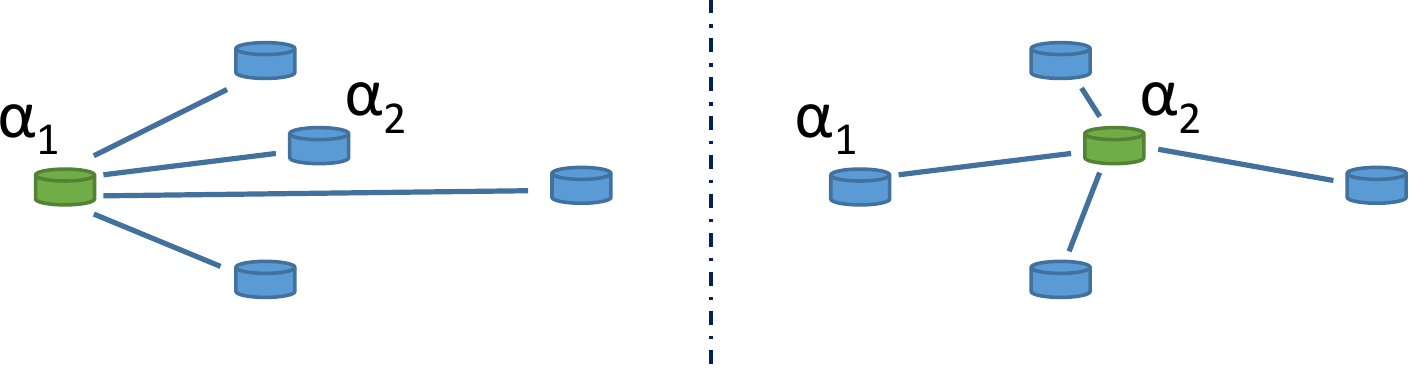}
    \caption{Cluster-head reassignment: Cluster-head $\alpha_1$ notices that it is not the best candidate anymore and passes the cluster token to $\alpha_2$, which is closer to the centroid of the cluster.}
    \label{fig:head_reassignment}
\end{figure}

Periodically or event-triggered, a cluster-head can decide to reassign its token to an idle cluster member, i.e., a member of the sub-tree that does not own another cluster token. For reassignment, the cluster-head selects the idle member that is closest to the centroid of the cluster. Triggering events are the despawning of the cluster-head or if a member is leaving the vicinity of the cluster-head. In the latter case, a cluster reassignment is only performed if the chosen idle member is closer to the centroid than the current cluster-head, while in the former case, it is always executed. Periodic reassignment can reduce communication link distances, but the token-passing overhead has to be considered. An example of the cluster-head reassignment in a level-1 cluster is depicted in Figure \ref{fig:head_reassignment}.

\subsubsection{Agent Transfer}
When agents leave the vicinity of their cluster head, moving closer to an adjacent cluster, an agent transfer can be initiated if a cluster-head reassignment was unsuccessful. In principle, this transfer could be conducted by evicting the agent from its original cluster, allowing it to join the closer cluster through the decentralized join process. However, this causes the agent to be unclustered for a brief period or, due to lack of coordination, for longer if the adjacent cluster altered unfavorably.

\begin{figure}
    \centering
    \includegraphics[width=\columnwidth]{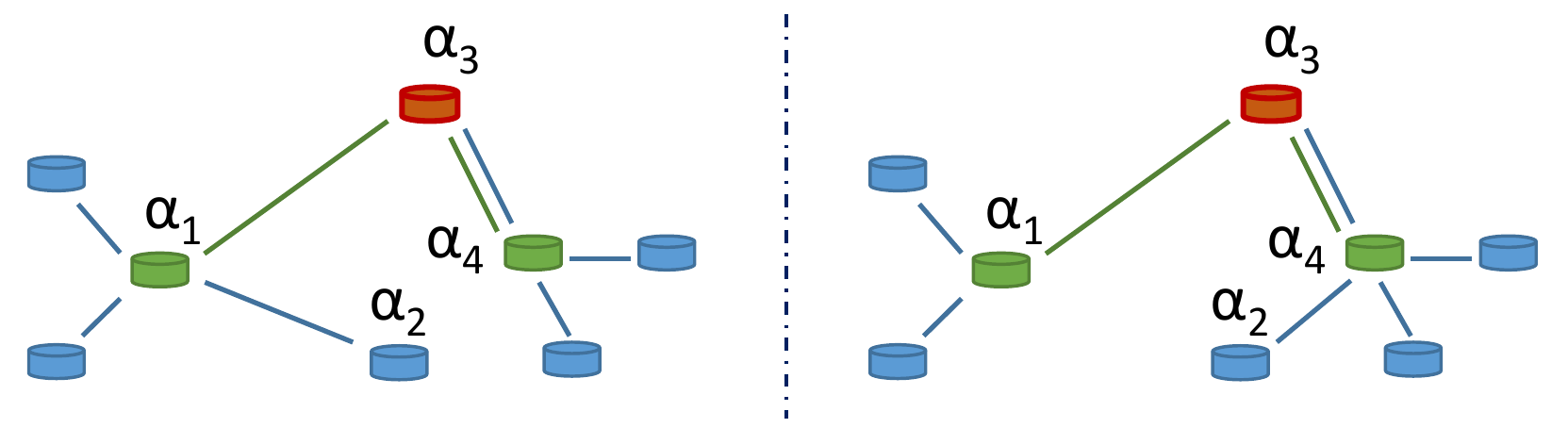}
    \caption{Transfer: $\alpha_1$ notices that $\alpha_2$ moved too far away and requests a transfer from $\alpha_3$. $\alpha_3$ evaluates if any of its members is better suited as cluster-head for $\alpha_2$. Since $\alpha_4$ is closer to $\alpha_2$ it transfers $\alpha_2$ to the cluster of $\alpha_4$.}
    \label{fig:transfer}
\end{figure}

A transfer protocol initiated by the cluster-head mitigates this risk through few direct messages between cluster-heads. The protocol is as follows: If a cluster-head registers that a member agent left its level-based vicinity, the cluster-head asks the head of its higher cluster if it has a member that is better suited as cluster-head for the agent in question. In such case, the agent is transferred to that cluster through the higher level cluster-head, adding coordination. An example of this transfer is shown in Figure \ref{fig:transfer}.

\begin{figure}
    \centering
    \includegraphics[width=\columnwidth]{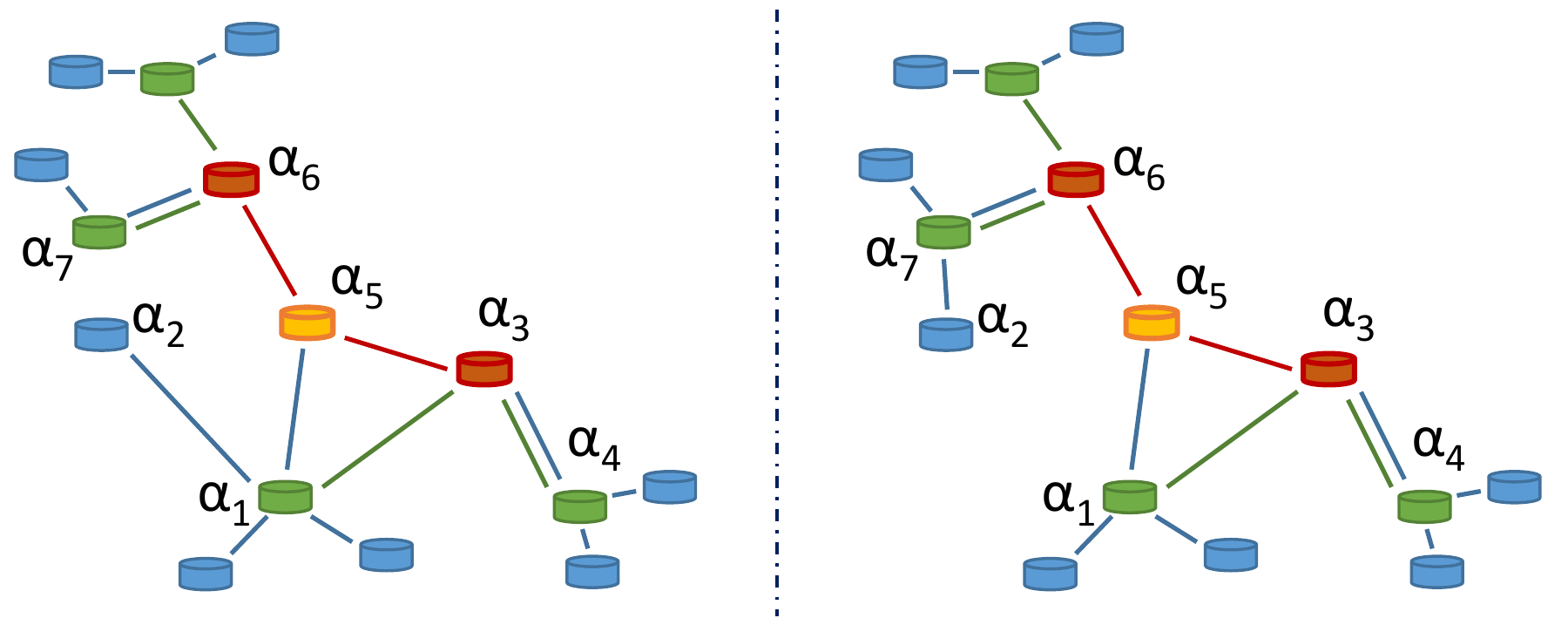}
    \caption{Transfer with one level of escalation: $\alpha_1$ notices that $\alpha_2$ moved too far away and requests a transfer from $\alpha_3$. $\alpha_3$ evaluates if any of its members is better suited as cluster-head for $\alpha_2$. Since $\alpha_4$ is not better suited, $\alpha_3$ escalates the request to $\alpha_5$. $\alpha_5$ asks all of its members if they have members that are better suited as cluster-head for $\alpha_2$. $\alpha_6$ responds that its member $\alpha_7$ is better, allowing $\alpha_5$ to initiate the transfer.}
    \label{fig:transfer_escalation}
\end{figure}

Occasionally, the best-suited cluster for the leaving agent is not a member of the higher level cluster-head but somewhere farther in the tree. An example of this is shown in Figure \ref{fig:transfer_escalation}. In that case, the transfer request can be escalated to the even higher cluster-heads. Each level of escalation, however, adds significant communication overhead. Therefore, we set the maximum escalation level as a hyperparameter, allowing trade-offs to be made based on the scenario at hand. In the example in Figure \ref{fig:transfer_escalation}, a maximum escalation level of one is sufficient. Note that, especially during transfer escalation, it has to be evaluated that \textit{Rule 3} is not violated through the transfer.

\subsubsection{Cluster Split}
When clusters are full, it is beneficial to split them up into two, such that the resulting clusters can assimilate incoming agents. To conduct the split, the cluster-head of the full cluster creates a new cluster, assigning an idle member closest to the cluster centroid as its cluster-head. The cluster members are then assigned to one of the two clusters according to their proximity.
\begin{figure}
    \centering
    \includegraphics[width=0.75\columnwidth]{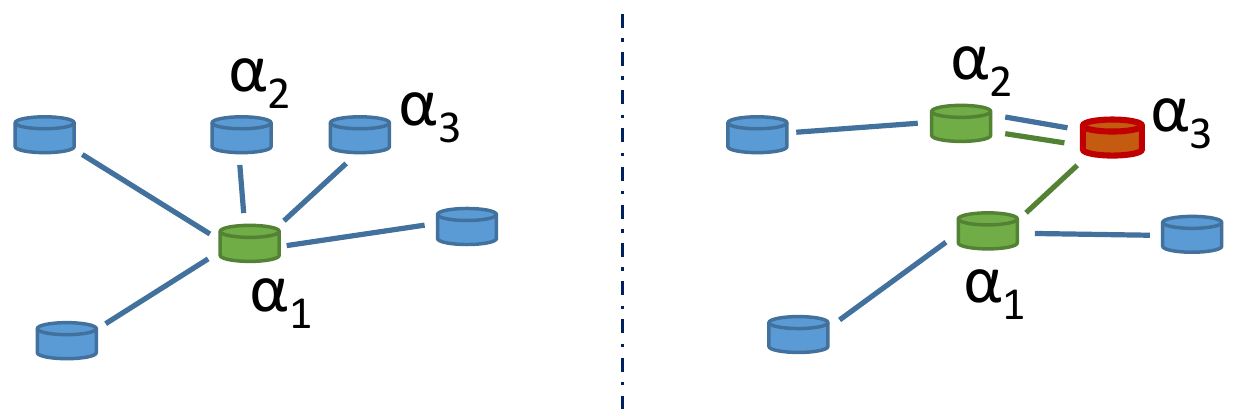}
    \caption{Split: With a maximum cluster size of 6 in this example, $\alpha_1$'s cluster is full. $\alpha_1$ initiates a split, selecting $\alpha_2$, which is closest to the centroid, as second cluster head. Agents are assigned to the cluster-head they are closest to. Since $\alpha_1$ was the boss, it selects $\alpha_3$ for creation of a level-2 cluster with $\alpha_1$ and $\alpha_2$ as members.}
    \label{fig:split}
\end{figure}
An example of this is shown in Figure \ref{fig:split}. The example further shows a special case in which the initiating cluster-head was the boss of the cluster tree. In this case, a split would create two disconnected sub-trees, which would need to use the decentralized communication method to find each other to form a higher-level cluster. To avoid this unnecessity, the split initiating cluster-head creates a higher-level cluster immediately, assigning one of its idle members as cluster-head, adding the two clusters resulting from the split into the member list.

\subsubsection{Cluster Assimilation}
\begin{figure}
    \centering
    \includegraphics[width=0.85\columnwidth]{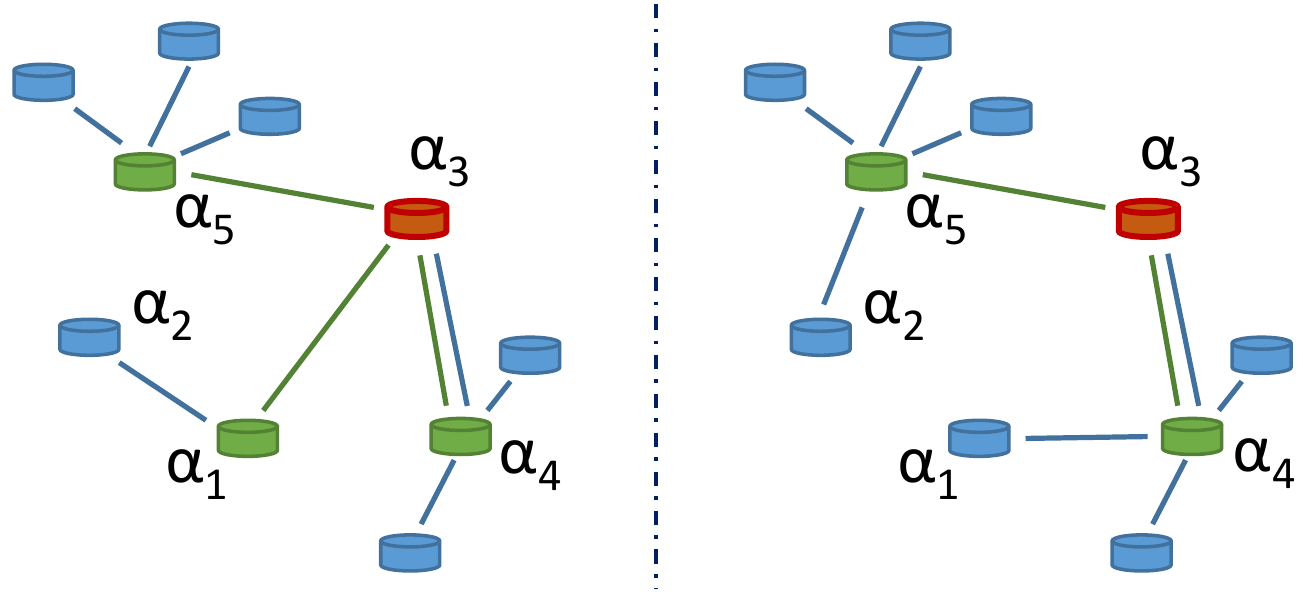}
    \caption{Assimilate: $\alpha_1$'s cluster only has two members, thus, $\alpha_1$ request from $\alpha_3$ to be assimilated into other clusters. $\alpha_3$ evaluates if $\alpha_4$ and $\alpha_5$ have enough room for $\alpha_1$ and $\alpha_2$. Since this is the case, $\alpha_3$ assigns $\alpha_1$ to $\alpha_4$ and $\alpha_2$ to $\alpha_5$.}
    \label{fig:assimilate}
\end{figure}

When agents are transferred, clusters can become very small, leading to unbalanced clustering. To mitigate this problem, small clusters can be assimilated by other clusters of the same level. An example is depicted in Figure \ref{fig:assimilate}. The cluster-head of the small cluster initiates the assimilation process. It requests assimilation from its cluster-head, passing a list of cluster members that need to be reassigned. The higher cluster-head evaluates whether its other members can accommodate these agents. To avoid clusters splitting immediately after, clusters can only accommodate agents if they are not full after assimilating. If all agents can be distributed, the transfers are conducted, and the cluster token of the assimilated cluster is destroyed. Cluster assimilation effectively regulates the width of the total cluster tree. The threshold on the number of agents, which, when subceeded, triggers an assimilation request is a hyperparameter, which is mostly dependent on the maximum cluster size.

\subsubsection{Boss Demotion}
\begin{figure}
    \centering
    \includegraphics[width=0.7\columnwidth]{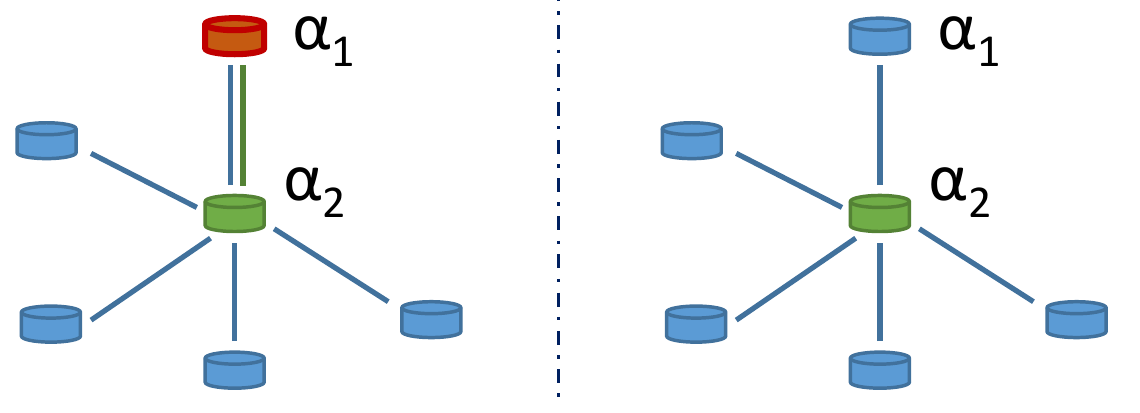}
    \caption{Boss demotion: $\alpha_1$ is the boss and only has one member. Therefore, it destroys its token, notifying $\alpha_2$ that it is the boss now.}
    \label{fig:demotion}
\end{figure}

When agents despawn and clusters are assimilated, the cluster-head of the highest existing cluster, the cluster boss, can have only one member. In that case, the boss does not improve the clustering and can thus destroy its cluster token. Consequently, its single member becomes the boss of the entire cluster tree. A simple example of a boss demotion can be seen in Figure \ref{fig:demotion}. Boss demotion regulates the depth of the cluster tree and is a crucial element for the case when all agents despawn.
\section{Observation Communication}
\label{sec:communication}
The previously described multi-level clustering is intended to be used as a backbone for communication within a multi-agent system. This section describes the propagation of a meta-belief, starting with definitions and operations.

\subsection{Meta Belief}
The belief of an agent is its estimate of the entire state of the system. We assume that the belief can be represented as a grid with cells containing localized information. To make the clustering communication application-agnostic, we define a meta belief that indicates the quality of information in each cell of the belief grid. An agent's total belief and an individual observation can be described with the meta-belief.

For cell $xy$ the meta belief $b_{xy} = (r_{xy}, a_{xy})$ consisting of a resolution $r_{xy}\in[0, 1]$ and a non-negative data age $a_{xy}\in\mathbb{R}^+_0$. The resolution indicates the spatial resolution of data within the cell and the data age specifies when the data in the cell was collected. The total meta belief for an entire grid of shape $m\times n$ can then be described through $B \in\mathcal{B}$ with $B = (R, A)$, in which $r_{xy}$ and $a_{xy}$ are the elements of $R \in [0,1]^{m\times n}$ and $A \in {\mathbb{R}^+_0}^{m\times n}$, respectively. The meta belief thus describes the quality and age of the belief of every cell.

\subsection{Aggregation}
When agents send their belief or a current observation through the cluster tree, the data must be aggregated. Especially when agents collect information about the same cells at the same or different time points, the aggregation or filtering method has to be defined. For this, we define a general aggregation function $\Phi:\mathcal{B}\times\mathcal{B}\mapsto\mathcal{B}$ which maps two beliefs into one. The cell-wise operations needed are denoted through $\phi$. In the following sections this function is abbreviated as the $+$ operator, i.e., $B_i + B_j := \Phi(B_i, B_j)$, where the specific aggregation function can have one of the following implementations:

\subsubsection{Age prioritization}
Age prioritization is most applicable in scenarios where the detailed knowledge of a cell is unimportant, but its recentness is. Its cell-wise implementation is defined as follows
\begin{equation}
    \phi_A(b_{1}, b_{2}) 
    = 
    \begin{cases}
        b_{1}, & a_{1} < a_{2}\\
        b_{2}, & a_{1} > a_{2}\\
        \begin{cases}
            b_{1}, & r_{1} \geq r_{2}\\
            b_{2}, & r_{1} < r_{2}
        \end{cases}, & a_{1} = a_{2}
    \end{cases}
\end{equation}
In age prioritization the belief and meta belief of one of the two cells is selected, i.e., no filtering, averaging, or similar is conducted on the beliefs. The cell of meta belief 1 is selected if its data age is lower or if both ages are equal if its resolution is higher than the ones in meta belief 2. In case both are equal it is assumed that selecting either should yield the same result. In this work we use age prioritization as it is well suited for wild-fire front monitoring especially for potential field algorithms. However, we still briefly list other methods.

\subsubsection{Resolution prioritization}
Resolution prioritization is similar to age prioritization with the difference that resolution is evaluated first. It is defined through
\begin{equation}
    \phi_R(b_{1}, b_{2}) 
    = 
    \begin{cases}
        b_{1}, & r_{1} > r_{2}\\
        b_{2}, & r_{1} < r_{2}\\
        \begin{cases}
            b_{1}, & a_{1} \leq a_{2}\\
            b_{2}, & a_{1} > a_{2}
        \end{cases}, & r_{1} = r_{2}
    \end{cases}
\end{equation}
Resolution prioritization could be applicable in scenarios that are stationary. 

\subsubsection{Meta score}
Instead of favoring one over the other, a score based on age and resolution can be calculated. An aggregation function based on a score could look like
\begin{equation}
    \phi_S(b_{1}, b_{2}) 
    = 
    \begin{cases}
        b_{1}, & s(b_{1}) \geq s(b_{2})\\
        b_{2}, & s(b_{1}) < s(b_{2})
    \end{cases}
\end{equation}
with the scoring function $s\mapsto \mathbb{R}$. A possible scoring function could be a weighted sum of the resolution and the inverse age as
\begin{equation}
    s(b) = r+\frac{w}{a+1}
\end{equation}
with $w$ as weighting parameter.

\subsubsection{Scenario dependent}
Another aggregation method could be scenario or application-dependent. In wildfire monitoring, for example, if a cell contains fire in one of the beliefs, data age could be prioritized, while for a cell without fire resolution the one with the higher resolution can be selected. Furthermore, the beliefs for each cell could be merged in order to filter noise.

\subsection{Belief Data and Compression}
The amount of data contained in a belief is assumed to be directly proportional to the resolution of the data. We further assume that the meta-belief itself contains a negligible amount of data compared with the belief itself. Therefore, the data in a belief can be defined with $d:\mathcal{B}\mapsto\mathbb{R}$ as 
\begin{equation}
    d(B) = d_0 \sum_{\forall x,y} r_{xy}
\end{equation}
summing the resolution of each cell, multiplying with the data per full resolution cell $d_0$. In the following, we omit $d_0$ as we only compare beliefs with the same data per full resolution cell.

To reduce the amount of data that is being transmitted through the network beliefs can be compressed. For this we define the cell-wise data compression $B_c = B/c_d$ that applies $r_c = r/c_d$ on each cell of the belief $B\in\mathcal{B}$ with the compression factor $c_d\in\mathbb{R} \geq 1$. Consequently, the data of the compressed belief $B_c\in\mathcal{B}$ is $d(B_c) = d(B)/c_d$.

With the definitions of compression and aggregation it is clear that 
\begin{equation}
    B_i + B_i/c_d = B_i, \quad\forall c\geq 1
    \label{eq:cancelling}
\end{equation}
holds, as the age is not affected by compression and a higher resolution is always favored. Furthermore, we define the $-$ operator that, if subtracting $B_i$ from $B_j$, removes every cell from a belief $B_j$ for which $\phi(b_i, b_j)=b_i$. For the total belief the following is the consequence:
\begin{equation}
    B_i + B_j = B_i + (B_j - B_i)
\end{equation}
In this work, subtraction is not explicitly mentioned but it is applied in the code for evaluation. It is beneficial for saving data as
\begin{equation}
    d(B_j) \geq d(B_j - B_i)
\end{equation}
since $B_j$ contains fewer or equal cells after subtraction.

\subsection{Distribution}
Each agent within the clustering scenario has a unique ID, with the set of IDs given by $\mathcal{I}$. If the ID is not set or unknown it is set to 0. The agent tuple $\alpha_i \in \mathcal{A}$ for agent $i\in \mathcal{I}$, is defined through
\begin{equation}
    \alpha_i = (\underbrace{\vphantom{\{}\mathbf{p}_i, \mathbf{v}_i, l_i}_\text{Motion}, \underbrace{O_i, B_i\vphantom{\{}}_\text{Belief}, \underbrace{h_i\vphantom{\{}}_\text{Level-1}, \underbrace{hc_i, \{m_{i,k}\}}_\text{Cluster Token} )   
\end{equation}
with the agent position $\mathbf{p}_i\in\mathbb{R}^3$, agent velocity $\mathbf{v}_i\in\mathbb{R}^3$, and the remaining battery life $l_i\in \mathbb{R}$ defining the agent motion. The last ego observation $O_i\in\mathcal{B}$ and the total aggregated belief $B_i\in\mathcal{B}$ define the belief of the agent. The ID of the level-1 cluster-head that every cluster agent has is given through $h_i \in \mathcal{I}$. Consequently, the agent $\alpha_{h_i}$ is the level-1 cluster-head of $\alpha_i$.  If the agent is a cluster-head, $hc_i\in\mathcal{I}$ defines a potential higher-level cluster-head and the list of $m_{i,k}\in \mathcal{I}, \forall k \in [1,\dots,K_i] $ are the $K_i\in\mathbb{N}$ members of the cluster. These two elements define the cluster token, which can be passed to other agents if necessary.

\subsubsection{Ego Belief}
The local observation of an agent $i$, its ego belief, is given through $O_i$. The cells that fall within the field of view of the agent centered around its position $\mathbf{p}_i$ are set to $(r,a) = (1, 0)$, otherwise the cells are $(r,a) = (0, \infty)$. With the local observation the total belief of the agent $i$ can be updated with the measurement step according to
\begin{equation}
    B_i \leftarrow B_i + O_i \label{eq:measure}
\end{equation}
with a period of $t_O$.

\subsubsection{Cluster Belief}
To describe the belief that is propagated through the cluster hierarchy, two functions need to be defined. The first function $\lambda:\mathcal{I}\times \mathbb{N}\mapsto\mathcal{B}$ describes the total \textit{lower} belief of agent depending on a level. It is defined through
\begin{equation}
    \lambda(j, x) = 
    \begin{cases}
    \sum_{k=1}^{K_j}\lambda(m_{j,k}, x-1)/c_d, & x>0\\
    O_j, & x = 0
    \end{cases}
    \label{eq:lower_belief}
\end{equation}
The \textit{lower} belief of an agent is the aggregation sum of the \textit{lower} belief of its members, compressed by a data compression factor $c_d$. This recursive function ends when the level reaches 0, in which case the lower belief is the last observation $O_j$. The second function $\Lambda:\mathcal{I}\times \mathbb{N} \mapsto \mathcal{B}$ returns the aggregated belief of a cluster-head agent and is given through
\begin{equation}
    \Lambda(j, x) = 
    \begin{cases}
    \lambda(j,x)+\Lambda(hc_j, x+1), & u(j,x) = 1\\
    \lambda(j,x), & u(j,x) = 0
    \end{cases}
    \label{eq:agg_belief}
\end{equation}
The aggregated belief of an agent is either the aggregation of the lower belief and the aggregated belief of the higher level cluster or only the lower belief. The logical function $u:\mathcal{I}\times \mathbb{N}\mapsto \{0,1\}$ given through
\begin{equation}
    u(j,x) = ( hc_j\neq 0) \land (\Delta t_j \geq t_\Lambda c_t^{x+1})
\end{equation}
contains two aspects. The first part is true if there is a higher level cluster, i.e., it is false if $\alpha_j$ is the boss of the sub-tree. The second part contains the concept of time compression or subsampling. We define $\Delta t_j$ as the time since the last aggregated view of the higher level cluster was received and $t_\Lambda$ as the period with which a level-1 cluster-head sends its belief to its members. Then the time compression or subsampling factor $c_t$ reduces the number of updates coming from higher level clusters exponentially with cluster level. Effectively, $u()$ indicates whether an update from the higher level cluster is available or not.

When an agent is connected to a cluster, i.e., it has a level-1 cluster-head, it can periodically update its total belief according to
\begin{equation}
    B_i \leftarrow B_i + \Lambda(h_i, 1)\label{eq:comm_update}
\end{equation}
with the period $t_\Lambda$.

\subsection{Propagation Example}
\label{sec:propagation_example}

\begin{figure}
    \centering
    \includegraphics[width=0.8\columnwidth]{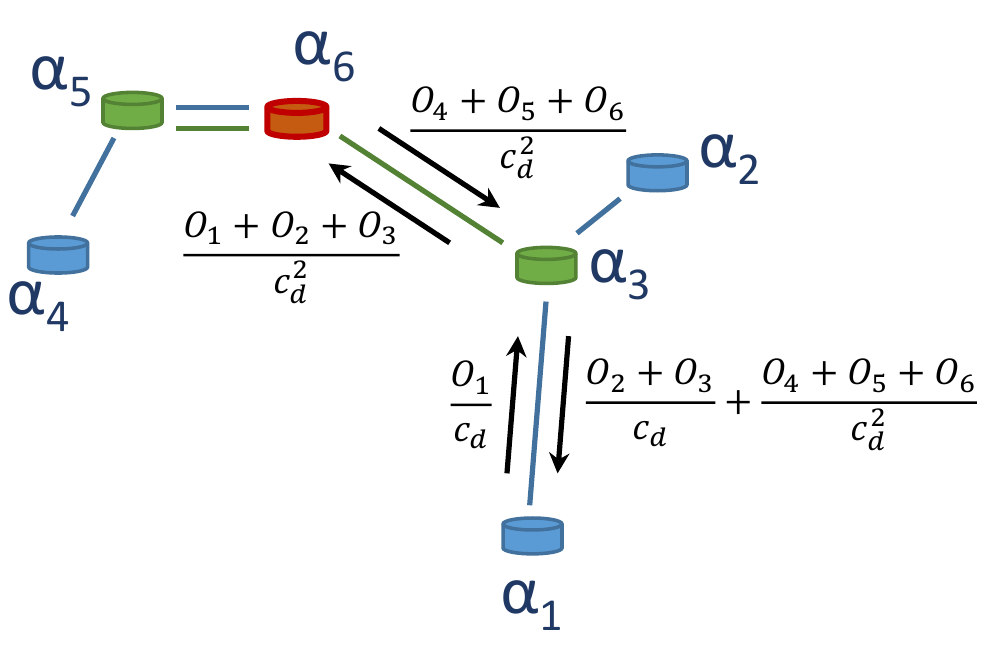}
    \caption{Belief update example showing a clustering example and observations being transferred for two exemplary links.}
    \label{fig:belief_example}
\end{figure}

To illustrate the belief update recursions, an example of a full belief update in the clustering scenario in Figure \ref{fig:belief_example} is given. For this example it is assumed that all update intervals are aligned, i.e., $u(j,x) = 1$ if $\alpha_j$ is not the boss.

An update containing a measurement and a cluster update can be written as 
\begin{equation}
    B_1 \leftarrow B_1 + B_{1,u}
\end{equation}
with the total belief update $B_{1,u}$ expanded through
\begin{subequations}
\begin{align}
    B_{1,u}&= O_1 + \Lambda(h_1,1) \label{eq:belief_ex_a} \\
    B_{1,u}&= O_1 + \lambda(3,1) + \Lambda(hc_3, 2) \label{eq:belief_ex_b} \\
    B_{1,u}&= O_1 + \lambda(3,1) + \lambda(6,2) \label{eq:belief_ex_c} \\
    B_{1,u}&= O_1 + \lambda(3,1) + \frac{\lambda(3,1)}{c_d} + \frac{\lambda(5,1)}{c_d}  \label{eq:belief_ex_d}\\
    B_{1,u}&= O_1 + \lambda(3,1) + \frac{\lambda(5,1)}{c_d}  \label{eq:belief_ex_e}\\
    B_{1,u}&= O_1 + \frac{\lambda(1,0)}{c_d} + \frac{\lambda(2,0)}{c_d} + \frac{\lambda(3,0)}{c_d} + \frac{\lambda(5,1)}{c_d}  \label{eq:belief_ex_f}\\
    B_{1,u}&= O_1 + \frac{O_1}{c_d} + \frac{O_2}{c_d} + \frac{O_3}{c_d} + \frac{\lambda(5,1)}{c_d}  \label{eq:belief_ex_g}\\
    B_{1,u}&= O_1 + \frac{O_2}{c_d} + \frac{O_3}{c_d} + \frac{\lambda(5,1)}{c_d}  \label{eq:belief_ex_h}\\
    B_{1,u}&= \underbrace{O_1\vphantom{\frac{O_2}{c_d}}}_{\text{ego}} + \underbrace{\frac{O_2}{c_d} + \frac{O_3}{c_d}}_{\text{level-1}} + \underbrace{\frac{O_4}{c_d^2} + \frac{O_5}{c_d^2} + \frac{O_6}{c_d^2}}_{\text{level-2}} \label{eq:belief_ex_i}
\end{align}
\end{subequations}
The recursion is unfolded as follows: In \eqref{eq:belief_ex_b} the aggregate belief definition in \eqref{eq:agg_belief} is used to unpack $\Lambda(h_1,1)$ and again in \eqref{eq:belief_ex_c} for $\Lambda(hc_3, 2)$. As $\alpha_6$ is the boss $\Lambda(6,2) = \lambda(6,2)$, which is then unfolded in \eqref{eq:belief_ex_d} according to the lower belief definition in \eqref{eq:lower_belief}. According to \eqref{eq:cancelling}, $\lambda(3,1) + \lambda(3,1)/c_d = \lambda(3,1)$, which is used in \eqref{eq:belief_ex_e} to simplify the term. In \eqref{eq:belief_ex_f}, $\lambda(3,1)$ is unpacked according to \eqref{eq:lower_belief} and recursion halts in \eqref{eq:belief_ex_g}. The term is again simplified in \eqref{eq:belief_ex_h} using \eqref{eq:cancelling}. Unpacking $\lambda(5,1)$ yields the final definition for the combined update in \eqref{eq:belief_ex_i}. In the fully expanded expression all parts of the update are visible: the uncompressed ego update, the once compressed level-1 update, and the twice compressed level-2 update.

Figure \ref{fig:belief_example} further illustrates the data that is effectively being transmitted between $\alpha_1$ and $\alpha_3$, and $\alpha_3$ and $\alpha_6$. Data is being compressed before sending up and reduced through cancellation when sending down.

\subsection{Base Station}
In a typical UAV scenario, a base station at the take-off and landing area is present. In this work, we utilize a base station to initialize the belief of a spawning agent. The base station's belief of the environment can be updated with the lower belief of the cluster boss periodically as
\begin{equation}
    B_{\text{base}} \leftarrow B_{\text{base}} + \lambda(j,x)/c_d
\end{equation}
with $\alpha_j$ being a level-x cluster-head and the boss. Furthermore, when agent $j$ lands, it can upload its entire belief to the base station resulting in the update
\begin{equation}
    B_{\text{base}} \leftarrow B_{\text{base}} + B_{i}
\end{equation}
The base station belief can then be used to initialize the belief of spawning agents. In the case that agent $j$ takes off, its belief can be assigned as
\begin{equation}
    B_{j} \leftarrow B_{\text{base}}
\end{equation}
The base station is essential if spawning agents need to react to previously collected data. Even if spawning agents directly connect to the cluster, they are only updated with the latest observations of the other agents. With the base station, they can directly explore according to a common belief. 

The base station could also be an explicit part of the clustering tree by always being the boss. This could bring many benefits, such as directly clustering spawning agents in a non-full level-1 cluster in the cluster tree and offloading computation and communication to an entity with unconstrained power. This work, however, focuses on clustering among homogeneous agents, where the base station is solely a support for initializing the belief of spawning agents. Therefore, we are not considering adding the base station to the clustering algorithm, but it will certainly be investigated in the future.

\section{Evaluation Methods}
\label{sec:eval_methods}

To evaluate the multi-level clustering, wildfire front monitoring scenarios are simulated, in which agents move according to different motion patterns, and are evaluated with metrics on cluster formation, monitoring performance, and communication links. The parameters that are used in the simulations are summarized in Table \ref{tab:parameters}.

\begin{table}
\center
\small
\begin{tabular*}{\columnwidth}{lc}
\toprule[1.5pt]
Parameter & Default Value\\
\midrule
Maximum cluster size & 8\\
Maintenance period [s] & 5 \\
Maximum transfer escalation levels & 1\\
Assimilation threshold & 2 \\
Level-1 distance threshold [m] & 100 \\
Data compression factor $c_d$ & 2 \\
Time compression factor $c_t$ & 2 \\
\midrule
Simulation time [s] & 2000 \\
Environment area [$\text{m}^2$] & $5000\times 5000$ \\
Maximum number of agents & 50 \\
UAV battery lifetime [s] & 500 \\
UAV spawning interval [s] & $[5, 15]$ \\
UAV spawning duration [s] & 1500 \\
\midrule
UAV velocity $v_0$ [m/s] & 20 \\
Repelling gain $w_r$ & $10^8$ \\
Age gain $w_a$ & 1 \\
Fire gain $w_f$ & 5000 \\
\midrule
Join period $P_j$ [s] & 1 \\
Create period $P_c$ [s] & 2 \\
Join/Create polling period [s] & 0.5\\
\midrule
Number of cells & $200 \times 200$ \\
Cell size $c_A$ [$\text{m}^2$] & $25\times 25$ \\
Fire update period [s] & 5 \\
Spreading factor $k_s$ & 0.02 \\
Burning rate $k_b$ & 0.02 \\
Spontaneous ignition probability $p_s$ & $10^{-7}$ \\
\bottomrule[1.5pt]\\
\end{tabular*}
\caption{Hyperparameters of MLC and their default value used in this work.}
\label{tab:parameters}
\vspace{5pt}
\end{table}

\subsection{Wildfire Front Monitoring}

We measure MLC's performance when applied in a wildfire front monitoring application. The wildfire simulation is similar to the one implemented for \cite{viseras2021wildfire}, with two differences. Firstly, we do not consider wind in this work, as its effect on the clustering is estimated to be marginal. Secondly, to make it necessary that some agents explore the entire environment, we start the fire at a random location and give a low probability for any cell to spontaneously ignite. This second difference makes the scenario significantly more complex and more realistic. It does not assume that the agents know a priori of the initial fire location or that the starting location is fixed and that the fire only spreads from that starting location. These assumptions are widely made in the literature \cite{viseras2021wildfire, Julian, Haksar2018}, but we think it makes the problem unrealistic.

The fire is spreading within a grid, in which any cell $(x,y)$ is either on fire or not denoted by $\beta_{xy} = 1$ or $\beta_{xy} = 0$, respectively. Each cell further has an amount of fuel $f_{xy}\in[0,1]$ that evolves according to 
\begin{equation}
    f_{xy} = f_{xy} - k_b \beta_{xy},
\end{equation}
burning away at a constant rate $k_b$ if the cell is on fire. The fire propagation is stochastic and the probability of any cell igniting is given through
\begin{equation}
    p_{xy} = 1 - \left( (1 - p_s)\left(1 - \prod_{\forall n,m} \frac{k_s\beta_{nm}}{|\mathbf{p}_{xy} - \mathbf{p}_{nm}|^2}\right)\right).
\end{equation}
The first part of the probability defined through $p_s$ is the probability that a cell spontaneously ignites. The second part is the probability that a cell in close vicinity propagates its fire to this cell. This part is parameterized through the spreading factor $k_s$ and decreasing with the squared distance between the cells' center points given through $\mathbf{p}_{xy}$ and $\mathbf{p}_{nm}$. To reduce computation complexity, only a small neighborhood is considered. A cell extinguishes and cannot ignite if its fuel $f_{xy}=0$.

When observed by the agents and described through a meta-belief, the resolution leads to an averaging of adjacent cells in the belief. The area over that is averaged is $c_A / r_{i, xy}$, with $c_A$ being the cell area and $r_{i, xy}$ the resolution of cell $(x,y)$ in the belief of the agent $\alpha_i$. Therefore, the belief of an $\alpha_i$ about the fire in cell $(x,y)$ is given through $\beta_{i,xy}\in[0,1]$, which is the amount of fire in the neighborhood of cell $(x,y)$ defined by the resolution.

\subsection{Motion Patterns}
In the evaluation, three different motion patterns are used that intend to yield different agent distribution throughout the environment. The patterns are random target tracking and data age minimization, and fire tracking using potential fields. In all scenarios, the agents spawn and despawn in the center of the environment. Agents spawn with specific battery life in random intervals for a fixed duration, after which agents only land. If a maximum number of agents is spawned, no new agent is spawned until an agent lands.

\subsubsection{Random Targets}
In the random target motion pattern, every agent randomly selects a target location within the environment perimeter, approaching it with constant velocity. When the target is reached, a new target is sampled until the agent's battery is nearly depleted. When the agent's battery is at a level that exactly suffices to fly back to the center of the environment, the target is changed to the center. The random target motion pattern creates independent motion but with a relatively high density in the center of the environment due to agents often flying from one side to the other.

\subsubsection{Potential Field}
The main motion pattern in this work is based on potential fields for path planning. The potential fields have three components: reciprocal repulsion, age attraction, and fire tracking. 

For reciprocal repulsion, agents are repelled from one another where the force is given through
\begin{equation}
    \mathbf{F}_r(\alpha_i) = w_r\sum_{\forall j\neq i} \frac{\mathbf{p}_i - \mathbf{p}_j}{|\mathbf{p}_i - \mathbf{p}_j|^3+1}
\end{equation}
with $w_r$ being a repelling weight. We assume that the position of all close agents are known. Further are the agents repelled by the perimeter of the environment in an equal manner.

Age attraction is added such that the agents explore the environment, venturing to cells that have not been updated for a while. The attractive force is given through
\begin{equation}
    \mathbf{F}_a(\alpha_i) = w_a \sum_{\forall x,y} a_{i,xy} \frac{\mathbf{p}_{xy} - \mathbf{p}_i}{|\mathbf{p}_{xy} - \mathbf{p}_i|^3+1}
\end{equation}
in which $a_{i,xy}$ is the data age of cell $(x,y)$ within the meta belief $B_i$ of agent $\alpha_i$. The location of the cell's center is given through $\mathbf{p}_{xy}$. The age attraction force is weighted with the factor $w_a$. 

Fire tracking is the only part of the potential field that is not based on the meta belief or the other agents and thus is the only component that is application dependent. The attractive force is calculated as follows:
\begin{equation}
    \mathbf{F}_f(\alpha_i) = w_f \sum_{\forall x,y} \beta_{i,xy} \frac{\mathbf{p}_{xy} - \mathbf{p}_i}{|\mathbf{p}_{xy} - \mathbf{p}_i|^2+1}
\end{equation}
in which $\beta_{i,xy}$ is the amount of fire in cell $(x,y)$ according to the belief $B_i$ of agent $\alpha_i$. With data compression this belief is averaged over adjacent cells according to the resolution in the meta belief. For fire attraction, the force is linear to the inverse of distance instead to its square, allowing agents to be attracted over farther distances.

The total force acting on each agent in the potential field is then given through
\begin{equation}
    \mathbf{F}(\alpha_i)  = \mathbf{F}_r(\alpha_i)  + \mathbf{F}_a(\alpha_i)  + \mathbf{F}_f(\alpha_i)
\end{equation}
and the resulting velocity of the agents is calculated as
\begin{equation}
    \mathbf{v}_i = v_0 \frac{\mathbf{F}(\alpha_i)}{|\mathbf{F}(\alpha_i)|}
\end{equation}
All agents are traveling at the same speed $v_0$ in the direction of the potential field.

With this potential field, agents can find and monitor a spreading wildfire. Together with the repelling term, the age attraction term makes the agents spread out quickly, allowing them to find potential fire quickly. When burning cells are found, the fire attraction term directs nearby agents towards the fire, while the repelling term makes them inspect different fire areas. After a while, the data age of distant cells is high enough such that the age attraction term pulls away agents on the perimeter of the fire, effectively making them look for new fire.

In the evaluation, two versions of this potential field method are used. The first is only using age attraction and reciprocal repulsion based on perfect belief, i.e., every agent communicating their full observation with every agent. The motion pattern is used to have exploring motion that is independent of the clustering algorithm. As the agents are not incentivized to be close to one another, this pattern yields the largest distances among agents. The second motion pattern uses all potential field terms based on the aggregated total belief of each agent. This pattern is used to evaluate how clustering parameters affect mission performance.

\subsection{Metrics}
We investigate three aspects: cluster formation, wildfire front monitoring performance, and the data links to evaluate the multi-level clustering. For each one, different metrics are used.

\subsubsection{Cluster Formation}
To evaluate how well the clusters formed, the \textit{main cluster ratio} is investigated throughout the scenarios. The main cluster ratio indicates how many agents belong to the biggest multi-level cluster that formed. If all agents are clustered into one tree, the main cluster ratio is 1. If no agent is clustered, it is $1/N$ with $N$ being the number of agents.

\subsubsection{Wildfire Front Monitoring}
For performance in wildfire front monitoring we utilize the fire miss ratio from \cite{viseras2021wildfire}, without normalizing it to the entire scenario. The fire miss ratio is defined as 
\begin{equation}
    m = 1 - \frac{\sum_{\forall x}\sum_{\forall y} \beta_{\text{obs},xy}}{\sum_{\forall x}\sum_{\forall y}\beta_{xy}}
\end{equation}
in which $\beta_{\text{obs},xy}$ is according to the belief 
\begin{equation}
    B_\text{obs} = \sum_{\forall i}O_i
\end{equation}
which is the collective direct observation of all agents. Effectively the fire miss ratio indicates how much fire is not currently being monitored by any agent.

Since different scenarios with different fires lead to significantly different fire miss ratios, we normalize the miss ratio per scenario. For each fire seed, one episode with perfect direct communication is evaluated and used as the baseline. The normalized fire miss ratio of an episode with clustering is then given as 
\begin{equation}
    \hat{m} = \frac{m_\text{MLC}}{m_\text{perfect}}
\end{equation}
which is equal to 1 if the performance is equal and higher if the performance is worse. 

\subsubsection{Data Links}
One of the main contributions of this work is to show that the number of links and the data rate for this communication scheme are significantly better than peer-to-peer communication of all agents. Therefore, we investigate the number of links, link distances, and data rate per agent.

\section{Results}
\label{sec:results}
The results are grouped into three parts. The first part shows how the main cluster ratio and data links depend on the maximum cluster size. In the second part, the wildfire front monitoring capabilities of the potential field method are investigated depending on data and time compression. The last part focuses on the evolution of the metrics throughout fire monitoring scenarios.

\subsection{Maximum Cluster Size}

\begin{figure}
    \centering
    \includegraphics[width=0.8\columnwidth]{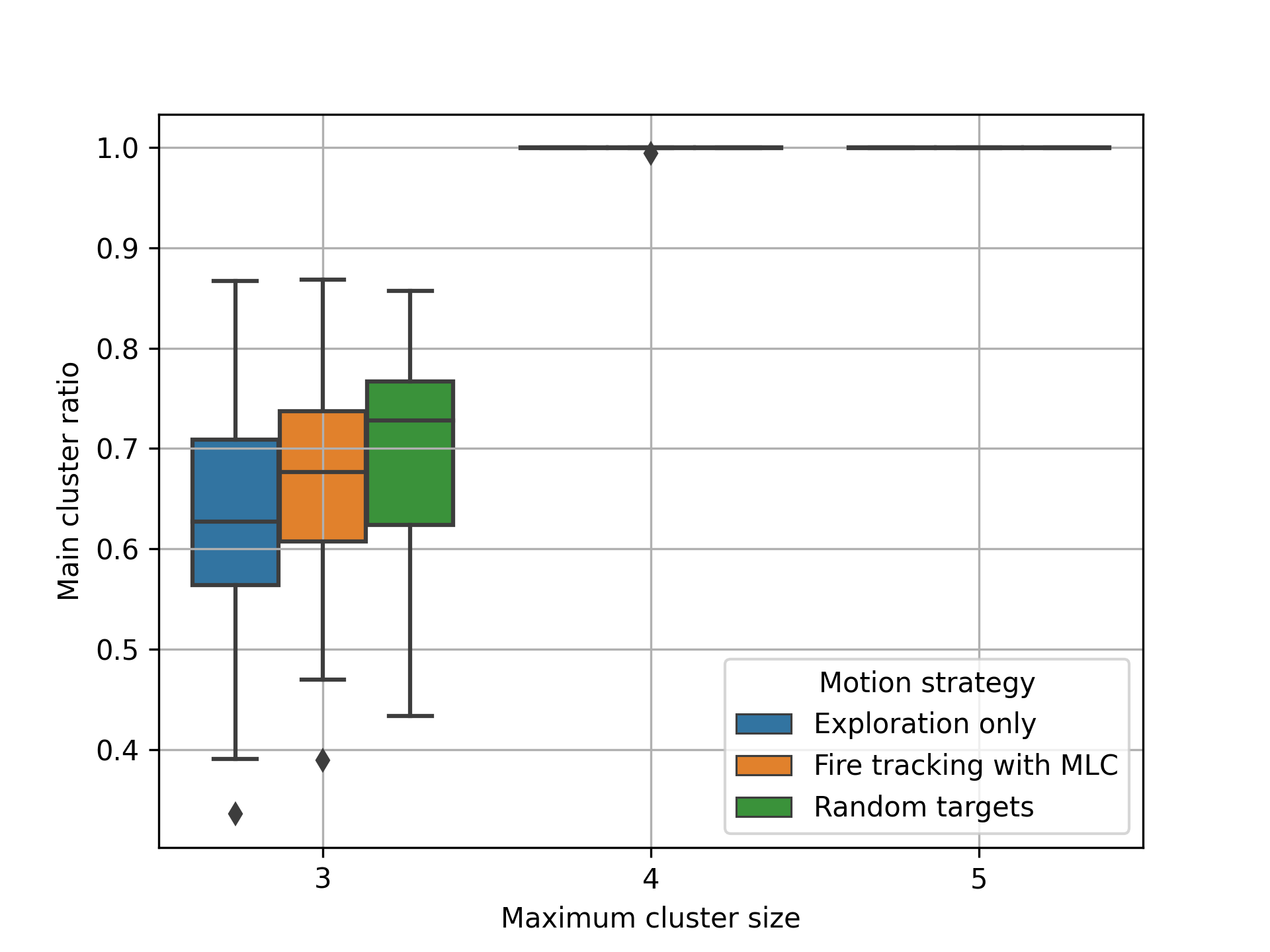}
    \caption{Box plot of the cluster formation depending on the maximum cluster size parameter for three different motion scenarios; for a maximum cluster size of 6-12 the main cluster ratio equals the values of cluster size 5.}
    \label{fig:stability}
\end{figure}

For this evaluation, 30 random fire scenarios are sampled, and agents are moving according to the three different motion strategies each, with maximum cluster sizes in $[3,12]$, yielding 900 simulations.
The first item to investigate is whether cluster formation is successful under the investigated parameter combinations. The main cluster ratio is shown in Figure \ref{fig:stability}. It can be seen that a maximum cluster size of 3 is not sufficient to form stable clusters, but everything above yields close to perfect clustering, i.e., every agent that spawns gets clustered to the main cluster tree within a few \textit{join} tasks. The result for a maximum cluster size of 3 is not surprising as the split implementation, for which full clusters are split, yields an effective maximum cluster size of 2. An arbitrary tree depth can only be created with only two agents per cluster if all clusters are exactly two members. Otherwise, no idle agent can function as the cluster head of a newly created highest level cluster. Because of that instability, a maximum cluster size of 3 is excluded in the following.

\begin{figure}
    \centering
    \includegraphics[width=\columnwidth]{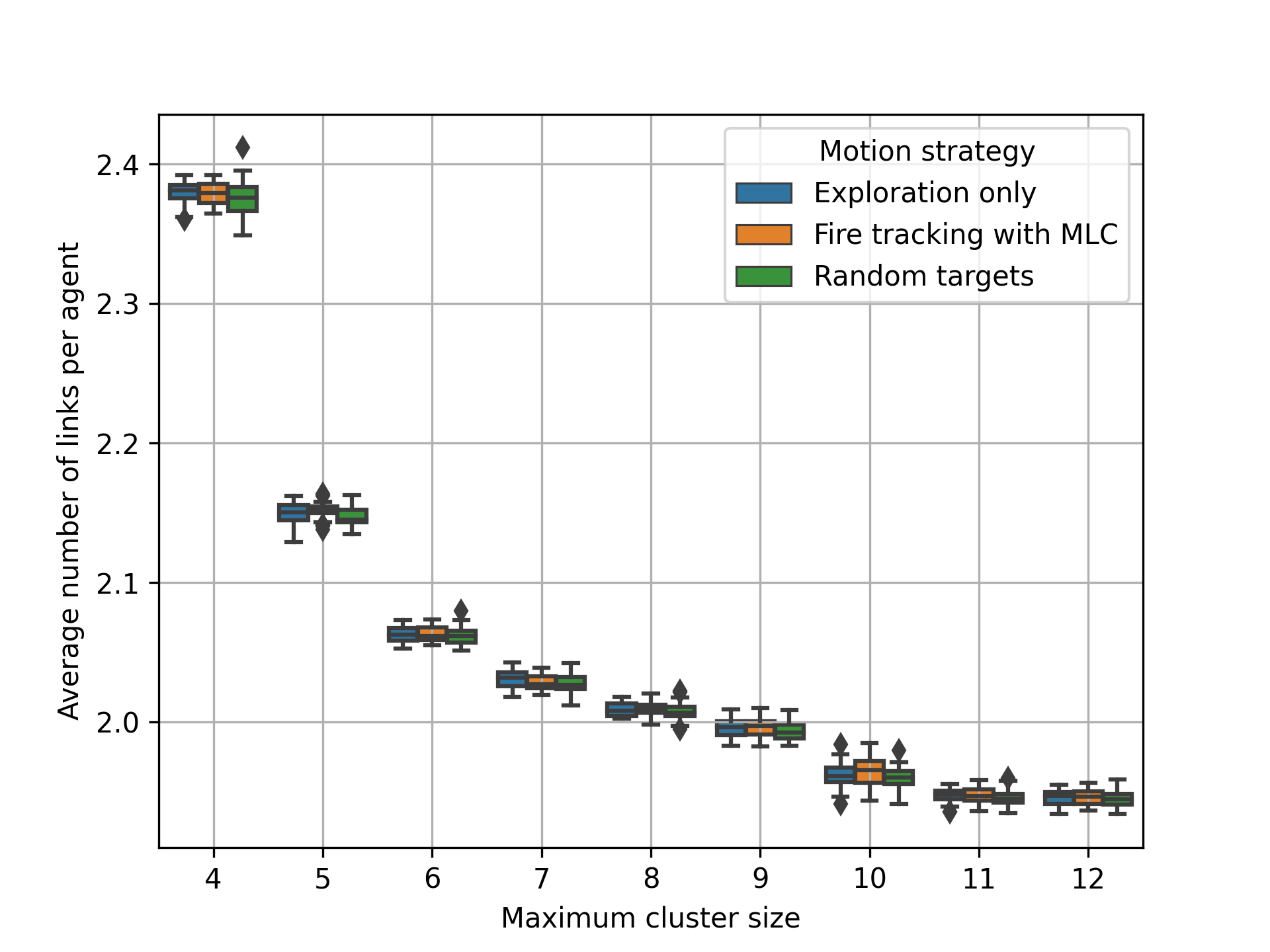}
    \caption{Average number of links per agent depending on the maximum cluster size and motion strategy.}
    \label{fig:links_number}
\end{figure}

The average number of links per agent depending on the maximum cluster size is depicted in Figure \ref{fig:links_number}. It can be seen that the number of links is independent of the motion strategy and decreases with the maximum cluster size. The total decrease throughout the plot is approximately 20\%. However, UAVs will need to be designed for the maximum number of links they could have to support: the cluster size plus two for the level-1 cluster-head and a higher level cluster or base station. Therefore, increasing the maximum cluster size can decrease the average number of links but increases the maximum number of links an agent can have.

\begin{figure}
    \centering
    \includegraphics[width=\columnwidth]{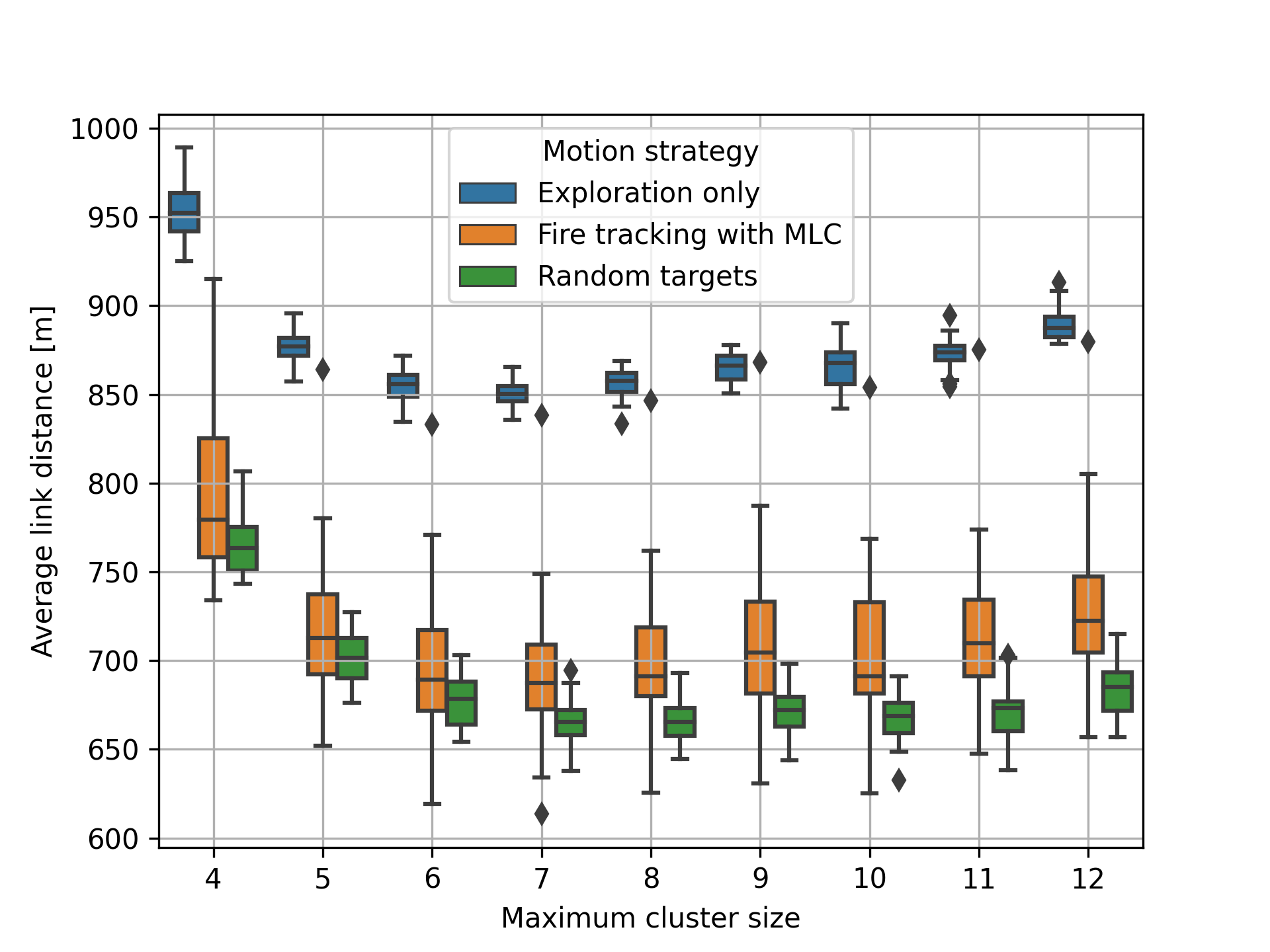}
    \caption{Average link distance, averaged over all links of all agents, depending on maximum cluster size and motion strategy.}
    \label{fig:links_distance}
\end{figure}

The average link distance is depicted in Figure \ref{fig:links_distance}. The first observation is that the link distance is strongly dependent on the motion strategy but follows the expected behavior. The agents are farthest apart for exploration only, and for random targets, they are closest together. The second observation is that with increasing cluster size, the link distance first decreases, up to a maximum cluster size of 7, from where it slightly increases again. A maximum cluster size of 7 or 8 appears to be the optimum for link distance under the given number of agents and given environment area.

\begin{figure}
    \centering
    \includegraphics[width=\columnwidth]{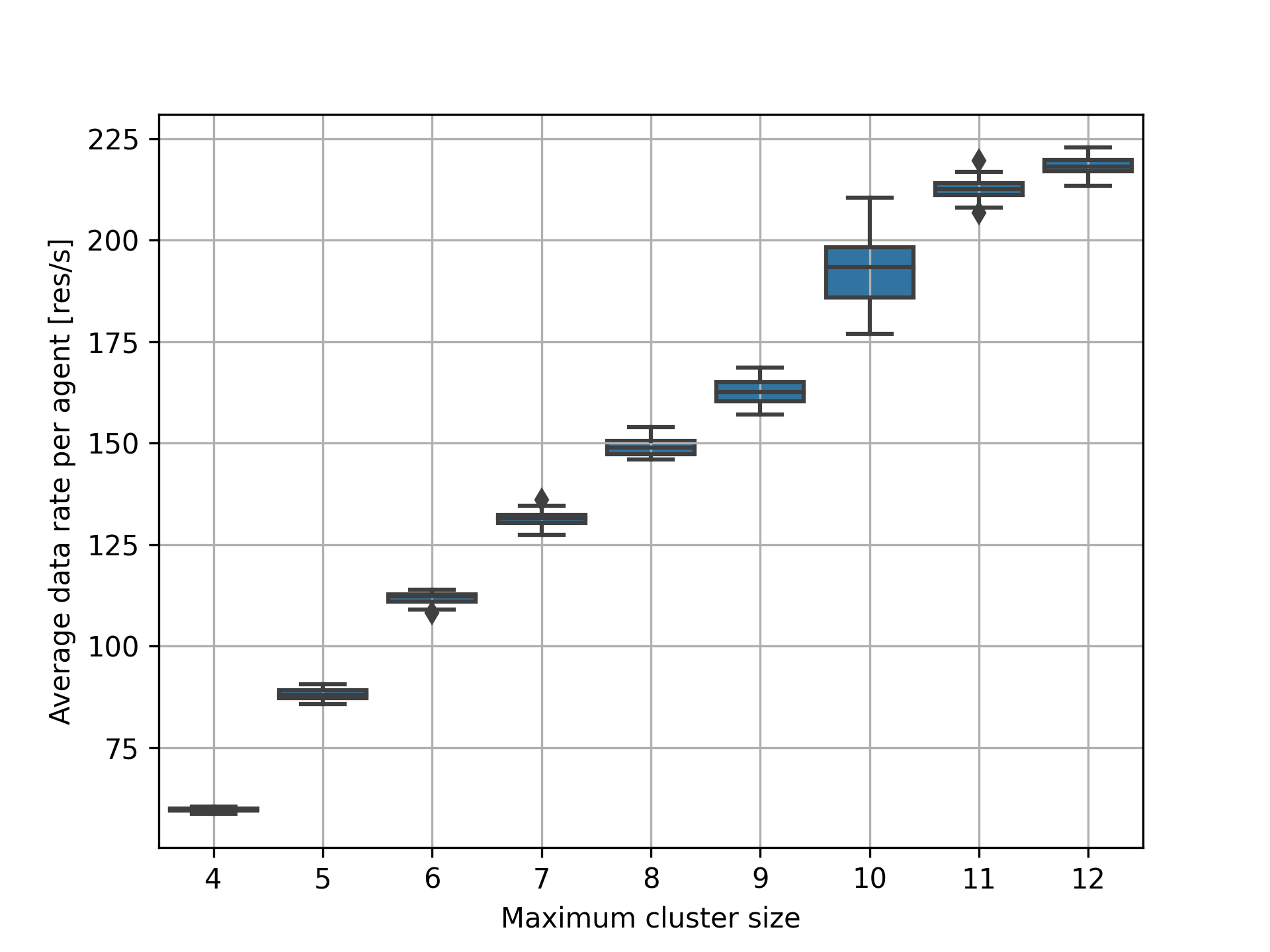}
    \caption{Average data rate per agent under fire tracking with MLC motion strategy depending on maximum cluster size.}
    \label{fig:data_rate_size}
\end{figure}

\begin{figure}
    \centering
    \includegraphics[width=\columnwidth]{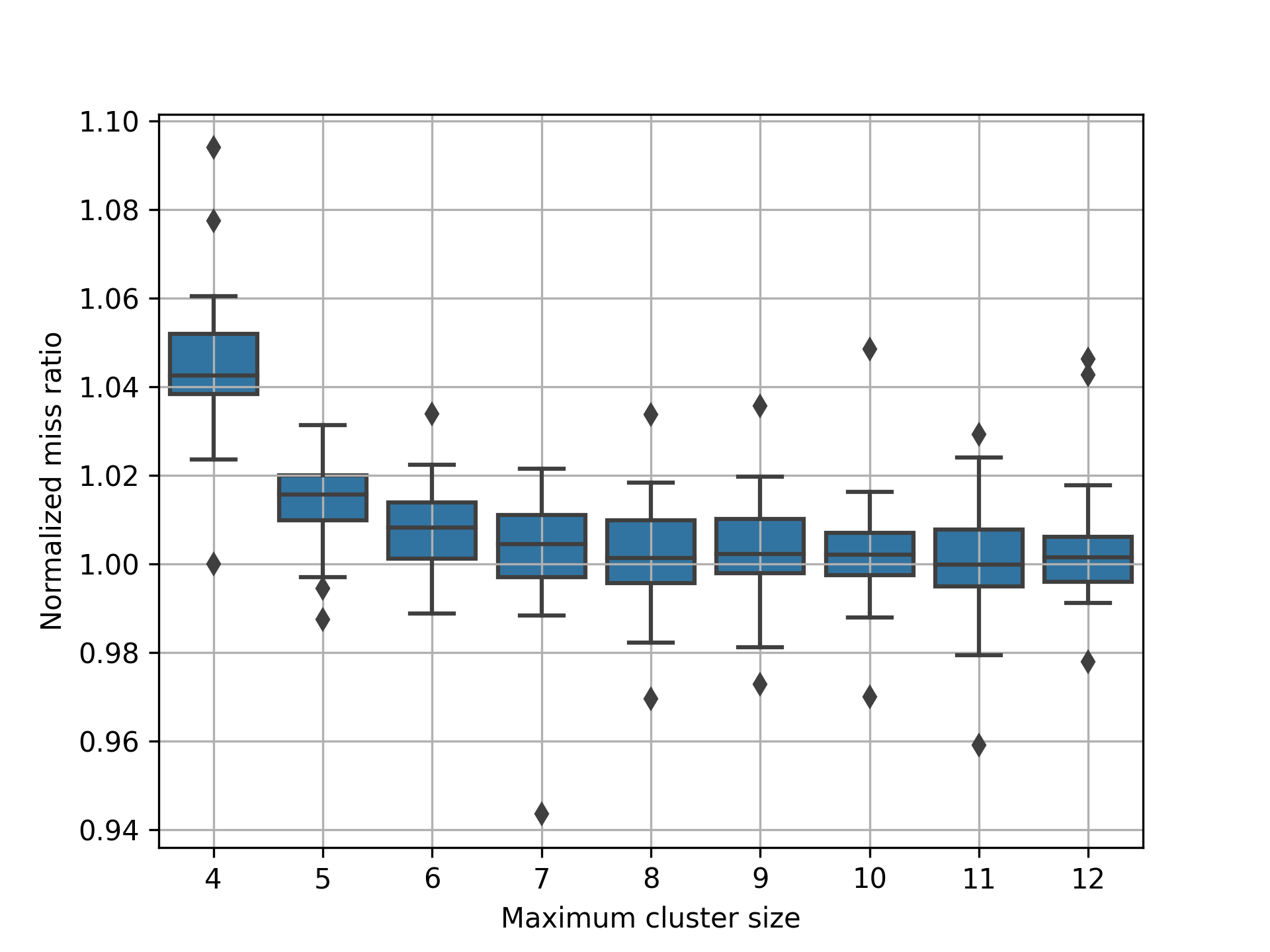}
    \caption{Normalized fire miss ratio under fire tracking with MLC motion strategy depending on maximum cluster size.}
    \label{fig:miss_ratio_size}
\end{figure}

The last graphics for this part of the results are shown in Figures \ref{fig:data_rate_size} and \ref{fig:miss_ratio_size}. The first shows the average data rate per agent using the fire tracking motion pattern with multi-level clustering for belief sharing, with $c_d=c_t=2$. The second shows the normalized fire miss ratio, which is better when lower, in the same scenarios. The data rate per agent increases roughly linearly with the maximum cluster size due to fewer data and time compression levels with bigger clusters. The normalized miss ratio decreases with the maximum cluster size saturating around a cluster size of 8.

The maximum cluster size is a hyper-parameter that has to be selected according to hardware constraints of the UAVs and the specific application. The results show that a lower limit of the maximum cluster size is four such that cluster formation is stable. An upper bound of this parameter is given by the constraints of the communication module on the UAVs, depending on the number of links that can be managed efficiently, which dictates the maximum cluster size as the number of maximum links minus two. The average number of links and the average link distance both benefit from an increase in the maximum cluster size where the latter saturates at a value of 7. The average data rate per agent is the only value that impedes the increase of the maximum cluster size to the maximum and thus should drive the selection process. Larger maximum cluster size is also beneficial for wildfire monitoring, while eight is sufficient to reach the best performance. Therefore, in the following, we use a maximum cluster size of 8.

\subsection{Wildfire Monitoring}

\begin{figure*}
\def\val{0.48}
    \centering
    \begin{subfigure}{\val\textwidth}
    \centering
    \includegraphics[width=\textwidth]{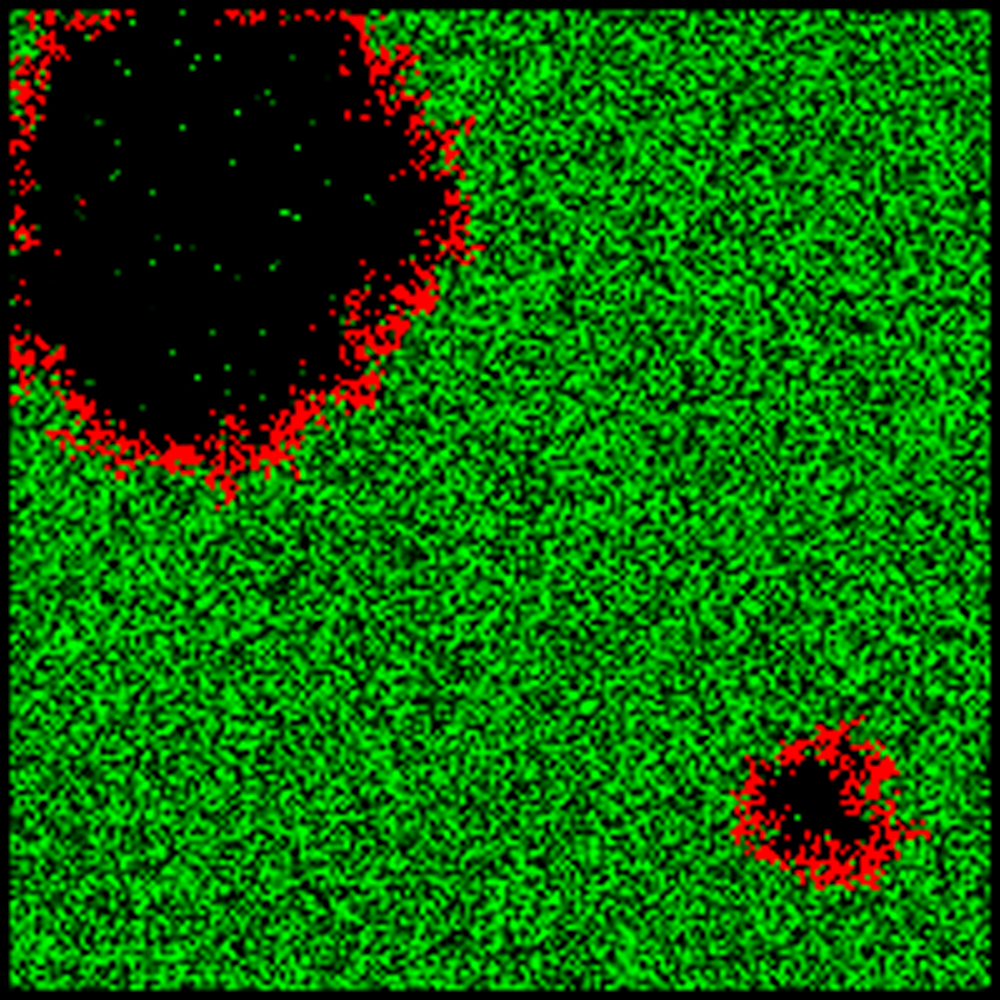}
    \caption{Ground truth}
    \label{fig:wildfire_ground_truth}
    \end{subfigure}
    \begin{subfigure}{\val\textwidth}
    \centering
    \includegraphics[width=\textwidth]{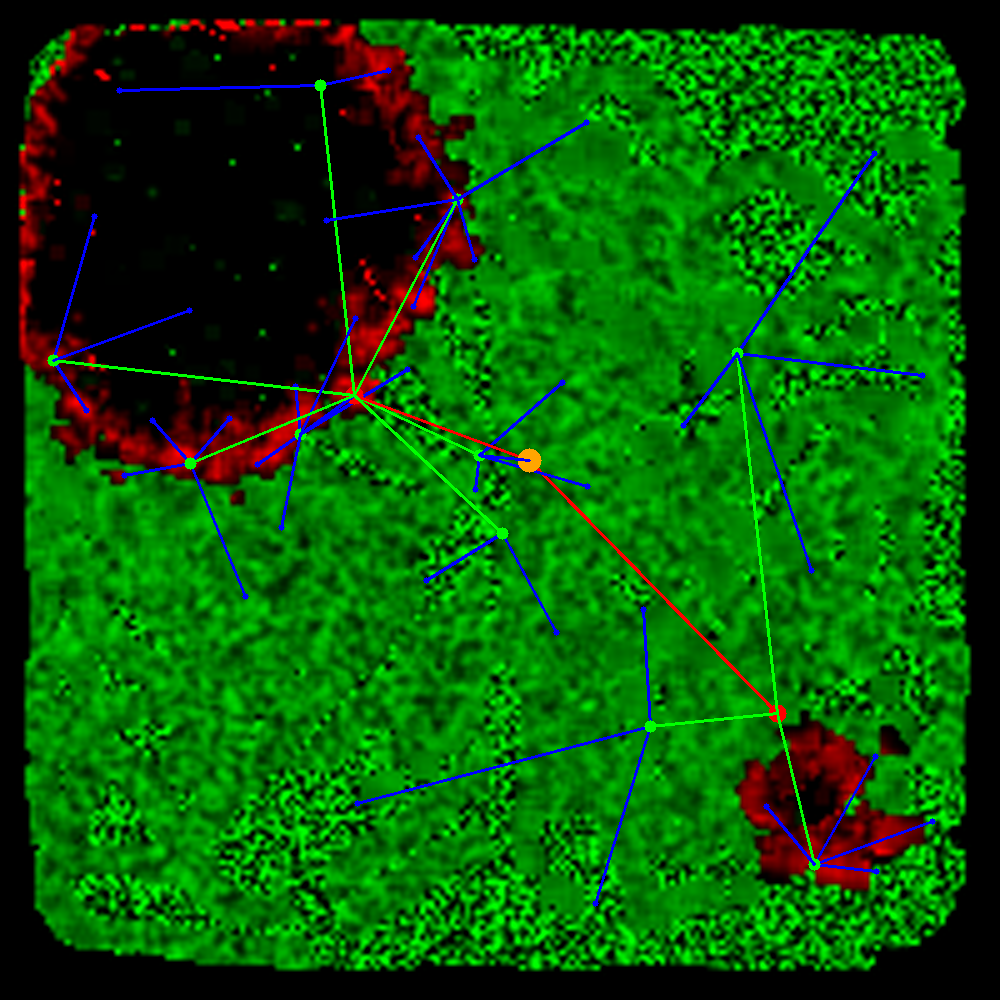}
    \caption{Base station full belief}
    \label{fig:wildfire_base}
    \end{subfigure}\hfill\\ \vspace{10pt}%
    \begin{subfigure}{\val\textwidth}
    \centering
    \includegraphics[width=\textwidth]{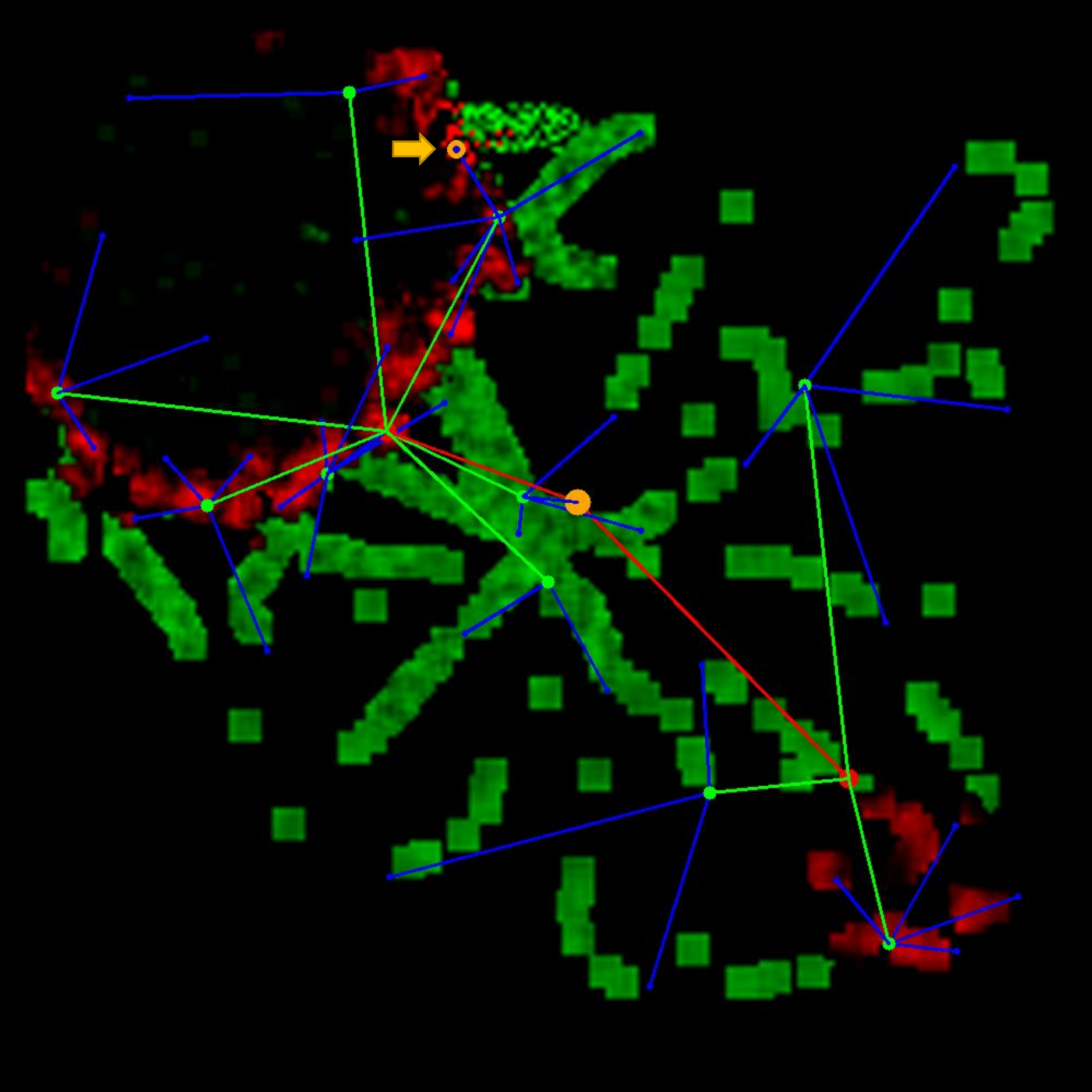}
    \caption{Top-left agent belief that is not older than 60 seconds}
    \label{fig:wildfire_agent_a}
    \end{subfigure}
    \begin{subfigure}{\val\textwidth}
    \centering
    \includegraphics[width=\textwidth]{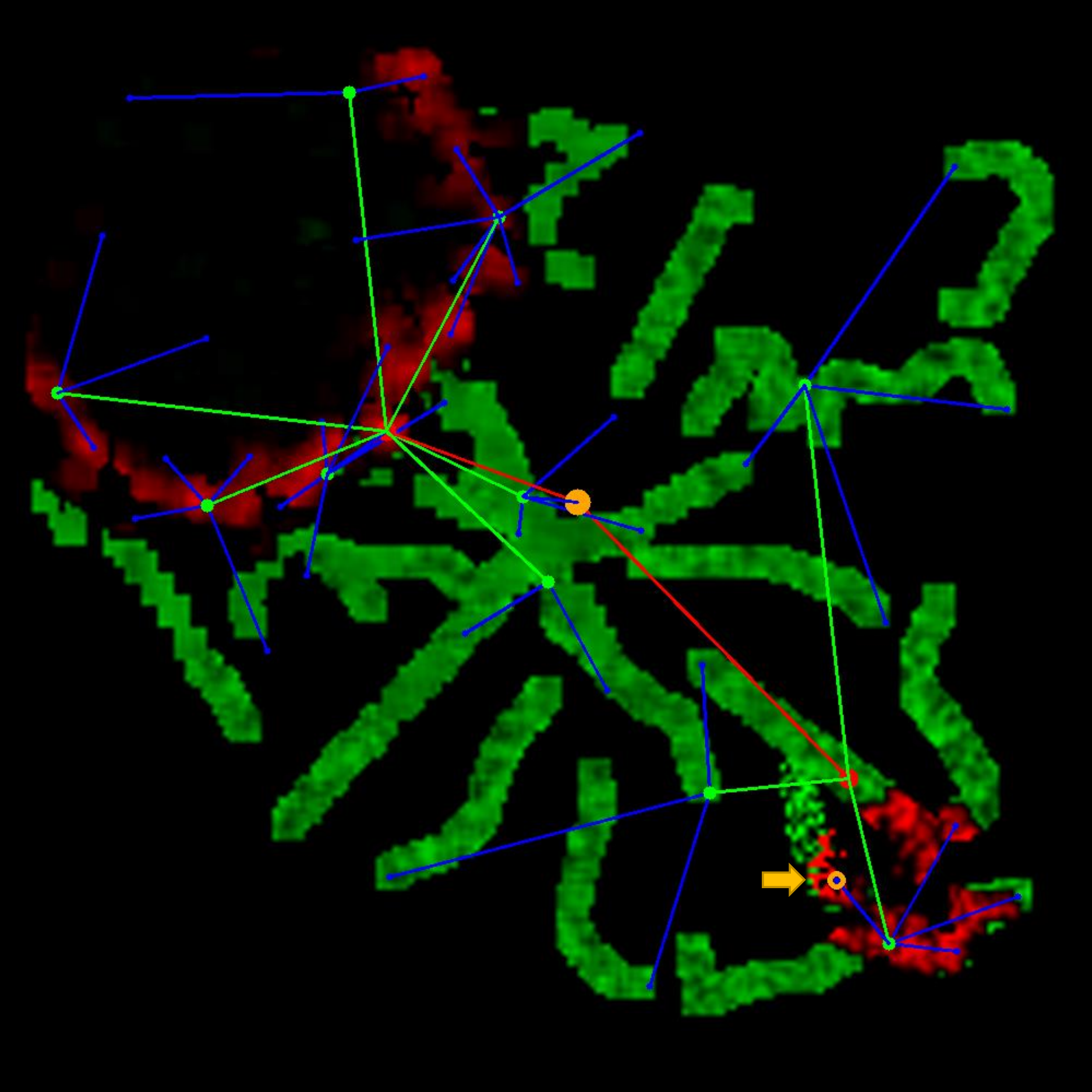}
    \caption{Bottom-right agent belief that is not older than 60 seconds}
    \label{fig:wildfire_agent_b}
    \end{subfigure}\hfill\\ \vspace{10pt}%
    \caption{Ground truth in (a) and different beliefs with blurriness depending on respective meta beliefs. Agents for the agent beliefs are highlighted with an orange circle and pointed at. The different resolution levels and levels of subsampling for distant beliefs are resulting from the data compression factor $c_d=3$ and time compression factor $c_t=2$ in this example.}
    \label{fig:wildfire_qualitative}
\end{figure*}

For the wildfire monitoring evaluation, we first give qualitative results to give an intuition of different beliefs in the system. Figure \ref{fig:wildfire_qualitative} shows a snapshot in time of a wildfire simulation from four different perspectives. The first graphic in \ref{fig:wildfire_ground_truth} shows the ground truth of the wildfire. The primary fire spreads in the top left while a smaller fire started on the bottom right. The second view in Figure \ref{fig:wildfire_base} shows the full belief of the base station that is accumulated from the boss updates and the belief of landing agents. The high-resolution patches, especially in the corners, result from landing agents that explored the outer perimeter uploading their belief. Since there is no fire in the bottom left and top right corner, agents will only travel there again when the data age of these cells is high enough. It can be seen that the total knowledge of the fire matches the ground truth very well, the only difference being a lower resolution. This shows qualitatively that the potential field method is working well in finding and tracking wildfires that are spreading according to the equations described in this work.

The Figures \ref{fig:wildfire_agent_a} and \ref{fig:wildfire_agent_b} show the belief of two agents that are as far apart in the cluster tree as possible. In these graphics, only the belief that is younger than 60 seconds is shown such that the differences are easier to perceive. The first agent is in the top-left monitoring a part of the fire there, and the second agent is monitoring the fire in the bottom-right. From the first agent's perspective in Figure \ref{fig:wildfire_agent_a} the resolution of the top left fire is high, and the frames blend smoothly into one another. The bottom-right fire, however, has very low resolution and is strongly subsampled. Nonetheless, it is still visible from the first agent's perspective that there is a fire in the bottom-right. In the belief of the second agent in \ref{fig:wildfire_agent_b} the perspective is reversed. It has a highly detailed belief of the bottom-right fire and a vague understanding of the top-left fire.

This qualitative result highlights the idea that is underlying the belief distribution in this work. Spatially close information is preserved and accurate, while distant information can be coarser and imprecise. 

\begin{figure}
    \centering
    \includegraphics[width=\columnwidth]{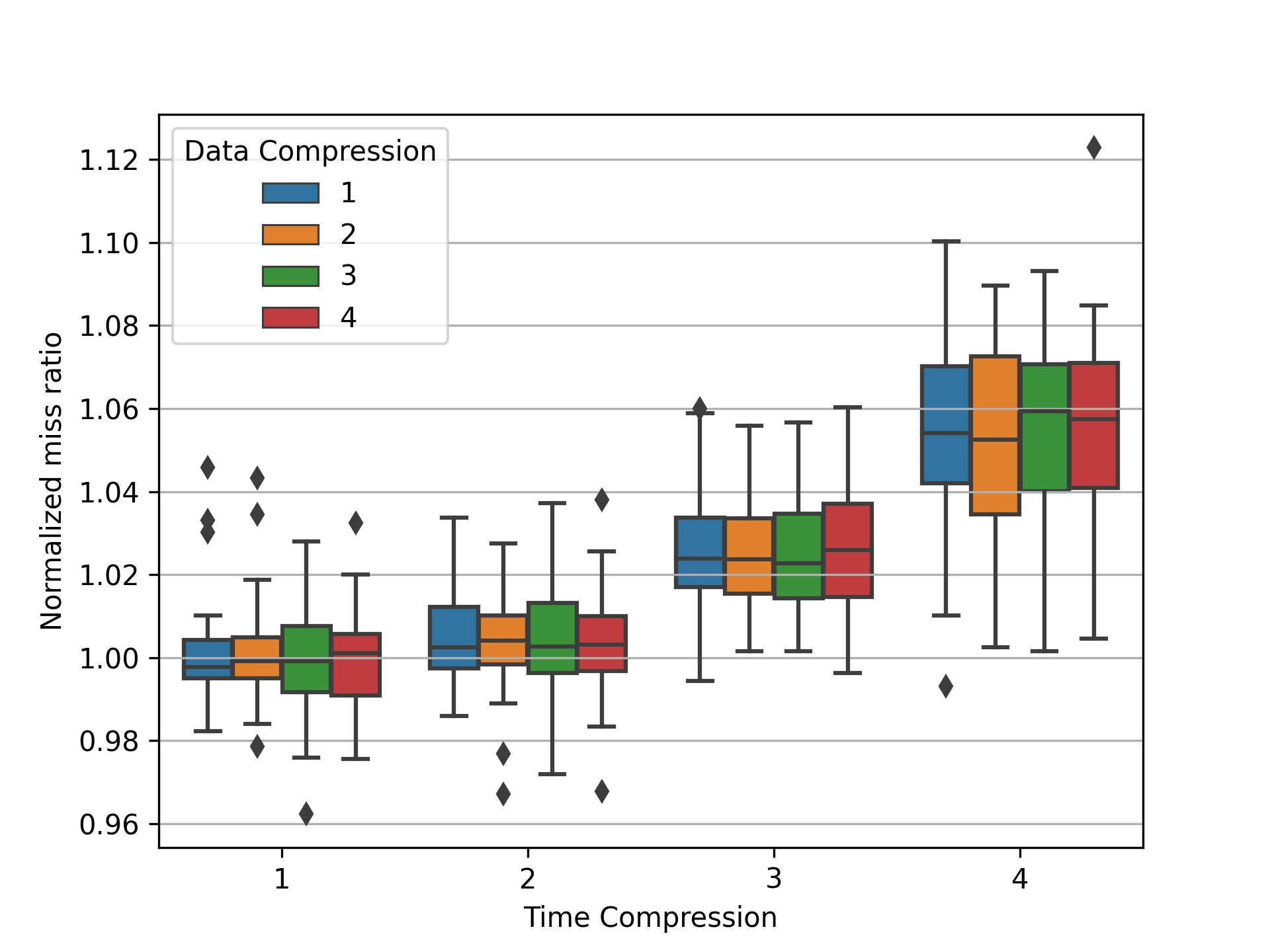}
    \caption{Normalized fire miss ratio depending on different levels of data compression and time compression. }
    \label{fig:fire_tracking}
\end{figure}

For a quantitative evaluation 30 wildfire scenarios were run with data compression $c_d\in[1,2,3,4]$ and time compression $c_t\in[1,2,3,4]$ yielding 480 simulations. The normalized fire miss ratio of the different combinations is depicted in Figure \ref{fig:fire_tracking}. It can be seen that data compression has little effect on the fire tracking performance. It might increase the variance in the result but not by a significant margin. Time compression, on the other hand, decreases performance significantly. To bring the results into perspective, the normalized miss ratio of exploration only motion and random motion are both 1.23, and the one of fire tracking potential fields without communication but with the base station for spawning and despawning is 1.18. Therefore, the fire tracking performance of any level of data and time compression is significantly better.

These results are to be expected. For potential fields, the exact location of the fire is not essential, as long as the amount of fire in the belief is preserved. Therefore, data compression does not change the fire tracking performance. With higher values of time compression, however, many frames of closer neighbors are missed, which changes the amount of fire in the belief. This significantly alters the potential field motion, apparently in a detrimental way. 

\begin{figure}
    \centering
    \includegraphics[width=\columnwidth]{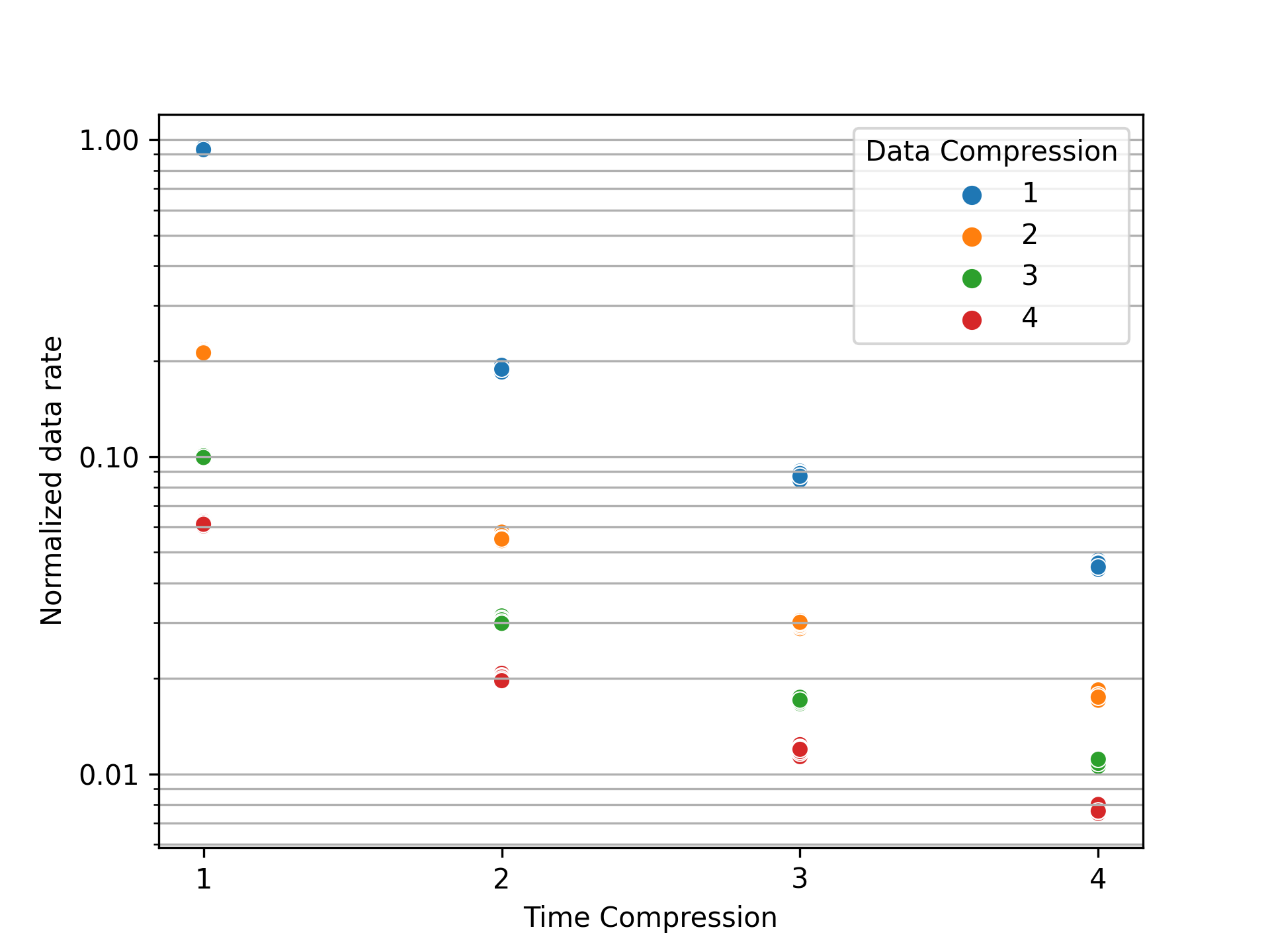}
    \caption{Average data rate normalized to direct communication depending on different levels of data compression and time compression.}
    \label{fig:data_rate}
\end{figure}

The average data rate that corresponds to the data and time compression combinations is shown in Figure \ref{fig:data_rate}. The data rate is normalized to direct uncompressed communication, which was used for the fire miss ratio normalization. As expected, the higher data and time compression levels drastically reduce the amount of data being transmitted. With no compression, the MLC communication is roughly on par with direct communication regarding the total data rate. With data and time compression set to 4, less than 1\% of the data has to be sent. With $c_d=c_t=2$, which is used in Figures \ref{fig:miss_ratio_size} and \ref{fig:miss_ratio_size} and in the following subsection, only roughly 5\% of the data is being transmitted.

These results show that MLC can reduce the data rate dramatically while keeping the performance high. The qualitative results showed that the developed clustering algorithm delivers the spatially dependent belief accuracy that was desired. Furthermore, the introduced compression parameters both have their advantages and disadvantages and allow the system designer to tune MLC for many applications. 

\subsection{Clustering Dynamics}

The evolution of the links within MLC throughout a wildfire simulation is evaluated in the following. For this, 30 fire simulations were run with potential field fire tracking motion and MLC. Every 2.5~s, the metrics are logged and displayed in time series. The metrics are collected for the MLC links and direct peer-to-peer links among all agents.

\begin{figure}
    \centering
    \includegraphics[width=0.9\columnwidth]{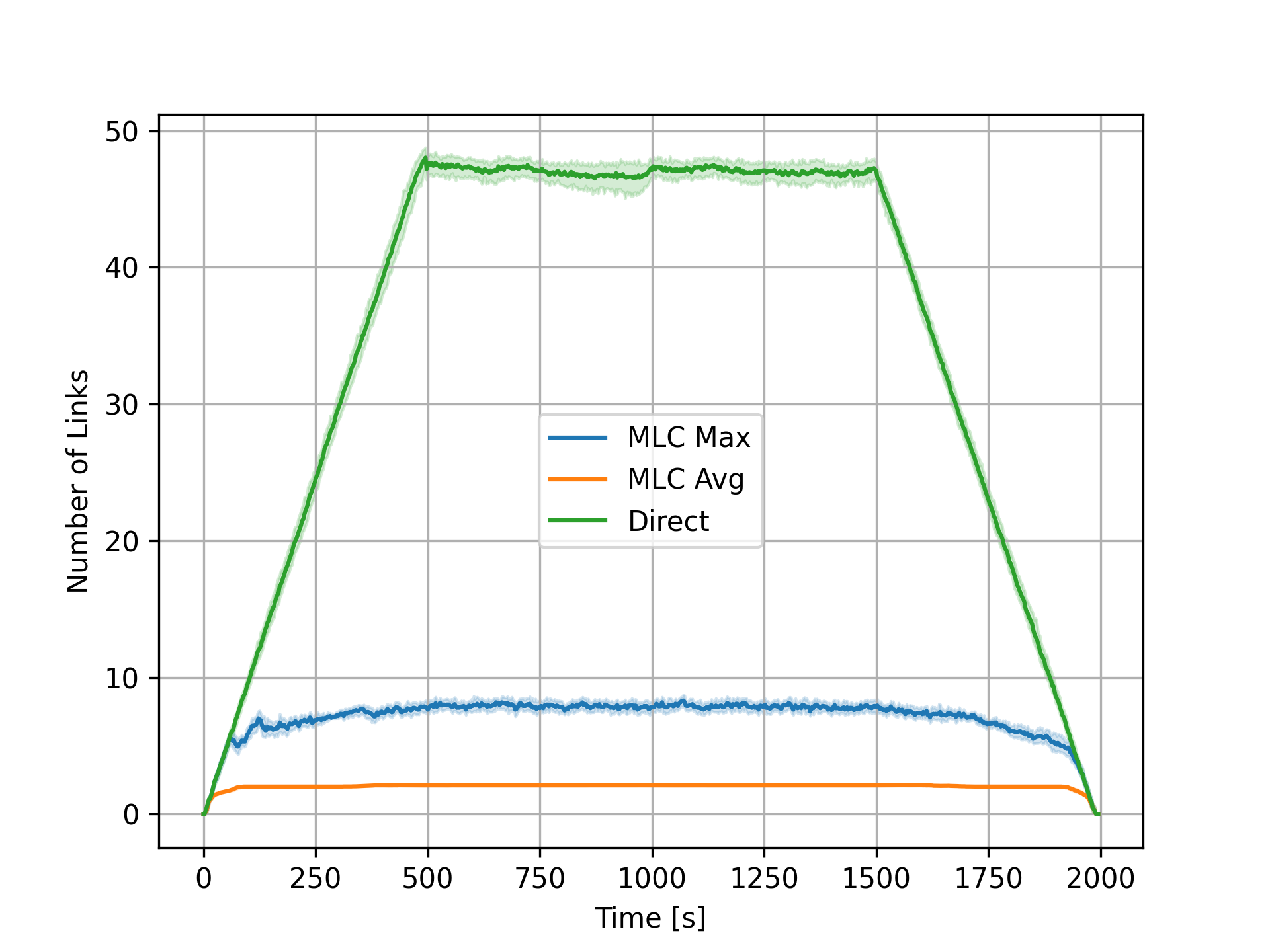}
    \caption{Time series showing the number of links for MLC and direct communication throughout multiple wildfire scenarios.}
    \label{fig:links_trace}
\end{figure}

The first time series in Figure \ref{fig:links_trace} shows the average and the maximum number of links in MLC with a maximum cluster size of 8, comparing with the number of links in direct communication. As seen in Figure \ref{fig:links_number}, the average number of links is around 2, and the maximum is slightly below the theoretical maximum of 10. The number of links in direct communication is equal to the number of agents minus one. It can be seen that MLC offers a significant benefit in the number of links in the worst case and average case throughout the entire mission. The average and maximum links stabilize quickly when the number of agents reaches its maximum, showing little to no variation.

\begin{figure}
    \centering
    \includegraphics[width=0.9\columnwidth]{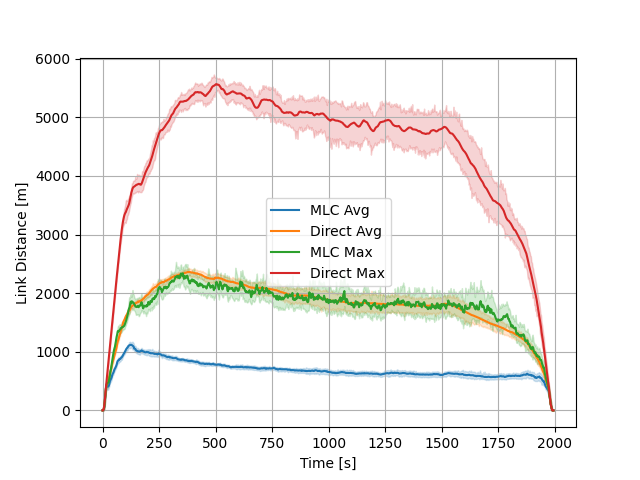}
    \caption{Time series showing the distance of links for MLC and direct communication throughout multiple wildfire scenarios.}
    \label{fig:distance_trace}
\end{figure}

Similarly, Figure \ref{fig:distance_trace} shows the average and maximum link distances of MLC and direct communication. Generally, the link distance first increases while the agents spread out, exploring the entire environment. When the environment was explored fully, and the fires are found, the distance decreases as agents gather at the fires. Some agents still explore, keeping the maximum link distances higher than the average. Comparing direct communication with MLC, both the average and the maximum link distances are by a factor of approximately 2.5 smaller for MLC. This shows that agents do not need to be in \textit{one-hop range} from each-other to form a stable clustering topology. The extend of the relaxation of this assumption will be investigated in future work. Interestingly the maximum distance in MLC tracks the average distance of direct communication. This is probably an artifact from selecting the close to optimal maximum cluster size parameter from Figure \ref{fig:links_distance} but will be further analyzed in future work.

\begin{figure}
    \centering
    \includegraphics[width=0.9\columnwidth]{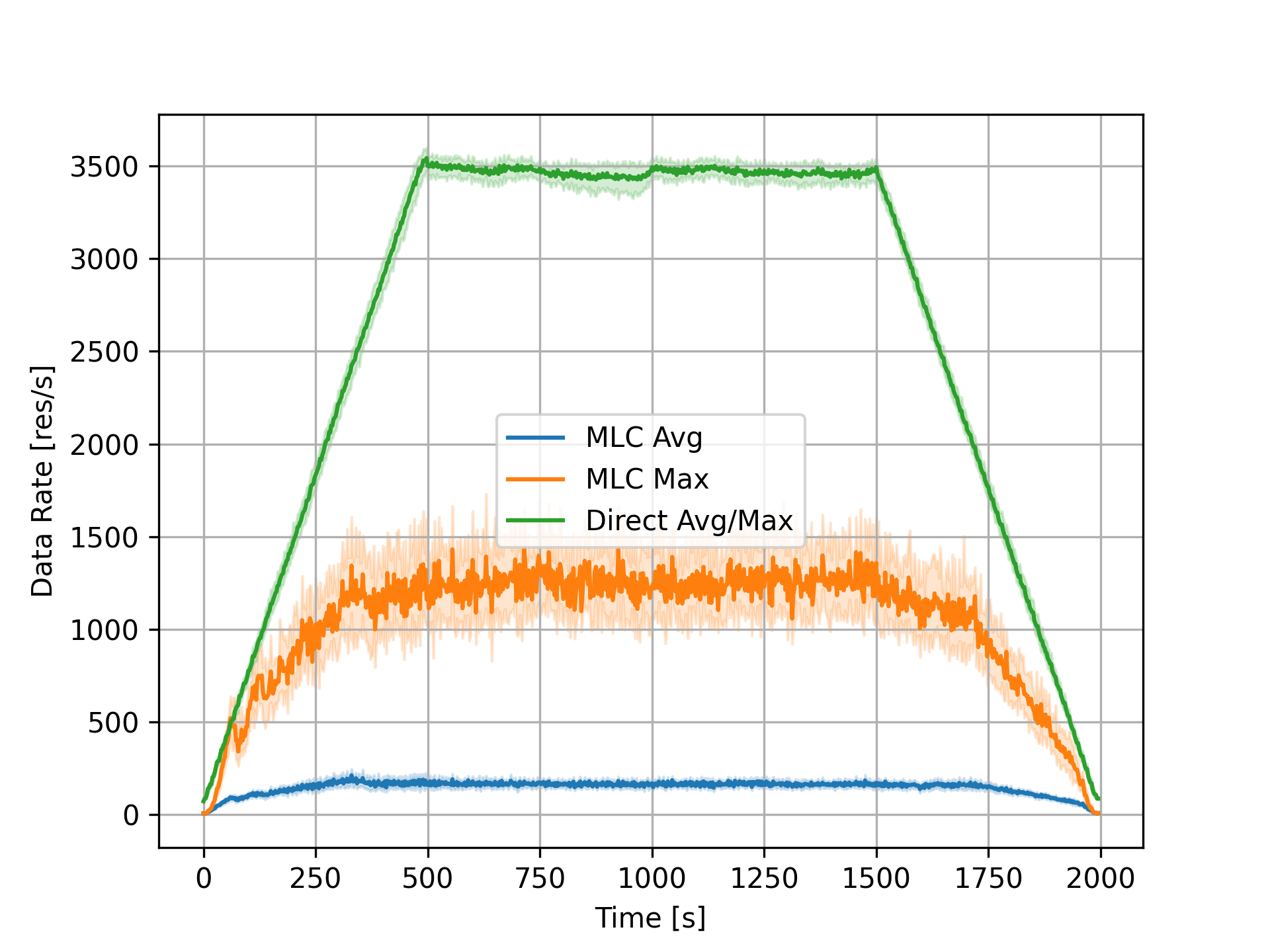}
    \caption{Time series showing the data rate per agent for MLC and direct communication throughout multiple wildfire scenarios.}
    \label{fig:rate_trace}
\end{figure}

The last graphic in Figure \ref{fig:rate_trace} shows the average and maximum data rate of the agents. Similar to the previous results, the average data rate of MLC is significantly lower than direct communication. The magnitude of the reduction is according to the results in Figure \ref{fig:data_rate} throughout the missions. As expected, the maximum data rate in MLC is significantly higher than the average since cluster-heads send more than idle agents, but still less than a third of the direct data rate.

The evaluation of the traces throughout the missions shows that MLC yields stable link numbers and distances and stable data rates. There are no unexpected outliers that could have been missed in the previous results, averaging over complete missions. The results show that MLC is applicable in scenarios that include spawning, despawning, and agent motion according to a global mission.
\section{Conclusion}
\label{sec:conclusion}
This work presented an autonomous hierarchical multi-level clustering (MLC) algorithm for multi-agent systems. Using MLC as a backbone for belief sharing among the agents allows for spatially oriented aggregation, compression, and dissemination of local observations yielding a total belief of the system for each agent. With extensive simulations for cluster formation stability and wildfire front monitoring, we showed the benefits of MLC. The wildfire tracking performance under the presented potential field method was not influenced by data compression within the cluster hierarchy. Time compression, i.e., time sampling, reduced the performance with high values of $c_t$. However, low values of $c_t$ reduced the data rate significantly, having only marginal effects on tracking performance. Evaluation of scenario traces further showed the stability of MLC in agile multi-agent systems, showing bounded values for the number of links, and significantly reduced link distances and data rates when compared with peer-to-peer communication.

For future work, MLC opens up three research branches. In the first branch, we will investigate MLC itself further, focusing on the induced communication overhead and link stability. Secondly, we will use MLC's belief sharing for multi-agent path planning with deep reinforcement learning. The spatially dependent resolution and freshness should be favorable for reinforcement learning as we previously showed in \cite{theile2020uav} for single agents and in \cite{bayerlein2021multi} for multi-agent systems. Using the presented potential fields method as a baseline, we will devise a deep reinforcement learning strategy for wildfire front monitoring. As a third branch, we aim to use the induced hierarchy of MLC and use it strategically in hierarchical planning methods such as hierarchical reinforcement learning, using MLC to solve the \textit{control problem} of multi-agent coordination.

\balance
\bibliographystyle{IEEEtran}
\bibliography{biblio}

\begin{thebibliography}{10}
\providecommand{\url}[1]{#1}
\csname url@samestyle\endcsname
\providecommand{\newblock}{\relax}
\providecommand{\bibinfo}[2]{#2}
\providecommand{\BIBentrySTDinterwordspacing}{\spaceskip=0pt\relax}
\providecommand{\BIBentryALTinterwordstretchfactor}{4}
\providecommand{\BIBentryALTinterwordspacing}{\spaceskip=\fontdimen2\font plus
\BIBentryALTinterwordstretchfactor\fontdimen3\font minus
  \fontdimen4\font\relax}
\providecommand{\BIBforeignlanguage}[2]{{%
\expandafter\ifx\csname l@#1\endcsname\relax
\typeout{** WARNING: IEEEtran.bst: No hyphenation pattern has been}%
\typeout{** loaded for the language `#1'. Using the pattern for}%
\typeout{** the default language instead.}%
\else
\language=\csname l@#1\endcsname
\fi
#2}}
\providecommand{\BIBdecl}{\relax}
\BIBdecl

\bibitem{Guerrero2013}
J.~Guerrero and Y.~Bestaoui, ``Uav path planning for structure inspection in
  windy environments,'' \emph{Journal of Intelligent and Robotic Systems},
  vol.~69, pp. 297--311, 01 2013.

\bibitem{Bouafif2018}
H.~Bouafif, F.~Kamoun, F.~Iqbal, and A.~Marrington, ``Drone forensics:
  Challenges and new insights,'' in \emph{2018 9th IFIP International
  Conference on New Technologies, Mobility and Security (NTMS)}, 2018, pp.
  1--6.

\bibitem{Kaleem2018}
Z.~Kaleem and M.~H. Rehmani, ``Amateur drone monitoring: State-of-the-art
  architectures, key enabling technologies, and future research directions,''
  \emph{IEEE Wireless Communications}, vol.~25, no.~2, pp. 150--159, 2018.

\bibitem{Ding2018}
G.~Ding, Q.~Wu, L.~Zhang, Y.~Lin, T.~A. Tsiftsis, and Y.-D. Yao, ``An amateur
  drone surveillance system based on the cognitive internet of things,''
  \emph{IEEE Communications Magazine}, vol.~56, no.~1, pp. 29--35, 2018.

\bibitem{Julian}
\BIBentryALTinterwordspacing
K.~D. Julian and M.~J. Kochenderfer, ``Distributed wildfire surveillance with
  autonomous aircraft using deep reinforcement learning,'' \emph{Journal of
  Guidance, Control, and Dynamics}, vol.~42, no.~8, pp. 1768--1778, 2019.
  [Online]. Available: \url{https://doi.org/10.2514/1.G004106}
\BIBentrySTDinterwordspacing

\bibitem{Haksar2018}
R.~N. {Haksar} and M.~{Schwager}, ``Distributed deep reinforcement learning for
  fighting forest fires with a network of aerial robots,'' in \emph{2018
  IEEE/RSJ International Conference on Intelligent Robots and Systems (IROS)},
  2018, pp. 1067--1074.

\bibitem{viseras2021wildfire}
A.~Viseras, M.~Meissner, and J.~Marchal, ``Wildfire front monitoring with
  multiple uavs using deep q-learning,'' \emph{IEEE Access}, 2021.

\bibitem{HASSANALIAN2018}
\BIBentryALTinterwordspacing
M.~Hassanalian, D.~Rice, and A.~Abdelkefi, ``Evolution of space drones for
  planetary exploration: A review,'' \emph{Progress in Aerospace Sciences},
  vol.~97, pp. 61--105, 2018. [Online]. Available:
  \url{https://www.sciencedirect.com/science/article/pii/S0376042117301884}
\BIBentrySTDinterwordspacing

\bibitem{Nuske2015}
S.~Nuske, S.~Choudhury, S.~Jain, A.~Chambers, L.~Yoder, S.~Scherer,
  L.~Chamberlain, H.~Cover, and S.~Singh, ``Autonomous exploration and motion
  planning for an unmanned aerial vehicle navigating rivers: Autonomous
  exploration and motion planning for a uav navigating rivers,'' \emph{Journal
  of Field Robotics}, vol.~32, 06 2015.

\bibitem{Meyer2015}
D.~Meyer, M.~Hess, E.~Lo, C.~E. Wittich, T.~C. Hutchinson, and F.~Kuester,
  ``Uav-based post disaster assessment of cultural heritage sites following the
  2014 south napa earthquake,'' in \emph{2015 Digital Heritage}, vol.~2, 2015,
  pp. 421--424.

\bibitem{Katila2017}
C.~J. Katila, A.~Di~Gianni, C.~Buratti, and R.~Verdone, ``Routing protocols for
  video surveillance drones in ieee 802.11s wireless mesh networks,'' in
  \emph{2017 European Conference on Networks and Communications (EuCNC)}, 2017,
  pp. 1--5.

\bibitem{GOODCHILD2018}
\BIBentryALTinterwordspacing
A.~Goodchild and J.~Toy, ``Delivery by drone: An evaluation of unmanned aerial
  vehicle technology in reducing co2 emissions in the delivery service
  industry,'' \emph{Transportation Research Part D: Transport and Environment},
  vol.~61, pp. 58--67, 2018, innovative Approaches to Improve the Environmental
  Performance of Supply Chains and Freight Transportation Systems. [Online].
  Available:
  \url{https://www.sciencedirect.com/science/article/pii/S136192091630133X}
\BIBentrySTDinterwordspacing

\bibitem{ponniah2021autonomous}
J.~Ponniah, M.~Theile, O.~Dantsker, and M.~Caccamo, ``Autonomous hierarchical
  multi-level clustering for multi-uav systems,'' in \emph{AIAA Scitech 2021
  Forum}, 2021, p. 0656.

\bibitem{McQuillan80}
J.~M. McQuillan, I.~Richer, and E.~C. Rosen, ``The new routing algorithm for
  the arpanet,'' \emph{IEEE Transactions on Communications}, vol.~28, no.~5,
  1980.

\bibitem{rfc1058}
\BIBentryALTinterwordspacing
{C. Hedrick}, ``{Routing Information Protocol},'' Internet Requests for
  Comments, {RFC Editor}, {RFC} 1058, 05 1988. [Online]. Available:
  \url{https://rfc-editor.org/rfc/rfc1058.txt}
\BIBentrySTDinterwordspacing

\bibitem{rfc2328}
\BIBentryALTinterwordspacing
J.~Moy, ``{OSPF Version 2},'' RFC 2328, Apr. 1998. [Online]. Available:
  \url{https://rfc-editor.org/rfc/rfc2328.txt}
\BIBentrySTDinterwordspacing

\bibitem{Dijkstra1959}
\BIBentryALTinterwordspacing
E.~W. Dijkstra, ``A note on two problems in connexion with graphs,''
  \emph{Numer. Math.}, vol.~1, no.~1, p. 269–271, Dec. 1959. [Online].
  Available: \url{https://doi.org/10.1007/BF01386390}
\BIBentrySTDinterwordspacing

\bibitem{Jacquet2001}
P.~Jacquet, P.~Muhlethaler, T.~Clausen, A.~Laouiti, A.~Qayyum, and L.~Viennot,
  ``Optimized link state routing protocol for ad hoc networks,'' in
  \emph{Proceedings. IEEE International Multi Topic Conference, 2001. IEEE
  INMIC 2001. Technology for the 21st Century.}, 2001, pp. 62--68.

\bibitem{rfc3626}
\BIBentryALTinterwordspacing
P.~Jacquet, ``{Optimized Link State Routing Protocol (OLSR)},'' Internet
  Requests for Comments, {RFC Editor}, {RFC} 3561, October 2003. [Online].
  Available: \url{https://www.rfc-editor.org/info/rfc3626}
\BIBentrySTDinterwordspacing

\bibitem{Perkins99}
C.~Perkins and E.~Royer, ``Ad-hoc on-demand distance vector routing,'' in
  \emph{Proceedings WMCSA'99. Second IEEE Workshop on Mobile Computing Systems
  and Applications}, 1999, pp. 90--100.

\bibitem{Perkins1994}
\BIBentryALTinterwordspacing
C.~E. Perkins and P.~Bhagwat, ``Highly dynamic destination-sequenced
  distance-vector routing (dsdv) for mobile computers,'' \emph{SIGCOMM Comput.
  Commun. Rev.}, vol.~24, no.~4, p. 234–244, Oct. 1994. [Online]. Available:
  \url{https://doi.org/10.1145/190809.190336}
\BIBentrySTDinterwordspacing

\bibitem{Kumar2000}
P.~Gupta and P.~Kumar, ``The capacity of wireless networks,'' \emph{IEEE
  Transactions on Information Theory}, vol.~46, no.~2, pp. 388--404, 2000.

\bibitem{Xie2004}
L.-L. Xie and P.~Kumar, ``A network information theory for wireless
  communication: scaling laws and optimal operation,'' \emph{IEEE Transactions
  on Information Theory}, vol.~50, no.~5, pp. 748--767, 2004.

\bibitem{Ozgur2007}
A.~Ozgur, O.~Leveque, and D.~N. Tse, ``Hierarchical cooperation achieves
  optimal capacity scaling in ad hoc networks,'' \emph{IEEE Transactions on
  Information Theory}, vol.~53, no.~10, pp. 3549--3572, 2007.

\bibitem{Ghaderi2009}
J.~Ghaderi, L.-L. Xie, and X.~Shen, ``Hierarchical cooperation in ad hoc
  networks: Optimal clustering and achievable throughput,'' \emph{IEEE
  Transactions on Information Theory}, vol.~55, no.~8, pp. 3425--3436, 2009.

\bibitem{Johnson96dynamicsource}
D.~B. Johnson and D.~A. Maltz, ``Dynamic source routing in ad hoc wireless
  networks,'' in \emph{Mobile Computing}.\hskip 1em plus 0.5em minus
  0.4em\relax Kluwer Academic Publishers, 1996, pp. 153--181.

\bibitem{Bellman}
\BIBentryALTinterwordspacing
R.~Bellman, ``A markovian decision process,'' \emph{Journal of Mathematics and
  Mechanics}, vol.~6, no.~5, pp. 679--684, 1957. [Online]. Available:
  \url{http://www.jstor.org/stable/24900506}
\BIBentrySTDinterwordspacing

\bibitem{sondik}
\BIBentryALTinterwordspacing
E.~J. Sondik, ``The optimal control of partially observable markov processes
  over the infinite horizon: Discounted costs,'' \emph{Operations Research},
  vol.~26, no.~2, pp. 282--304, 1978. [Online]. Available:
  \url{http://www.jstor.org/stable/169635}
\BIBentrySTDinterwordspacing

\bibitem{Kaelbling95planningand}
L.~P. Kaelbling, M.~L. Littman, and A.~R. Cassandra, ``Planning and acting in
  partially observable stochastic domains,'' \emph{Artificial Intelligence},
  vol. 101, pp. 99--134, 1995.

\bibitem{khatib1985}
O.~Khatib, ``Real-time obstacle avoidance for manipulators and mobile robots,''
  in \emph{Proceedings. 1985 IEEE International Conference on Robotics and
  Automation}, vol.~2, 1985, pp. 500--505.

\bibitem{Boutilier1996}
C.~Boutilier, ``Planning, learning and coordination in multiagent decision
  processes,'' in \emph{Proceedings of the 6th Conference on Theoretical
  Aspects of Rationality and Knowledge}, ser. TARK '96.\hskip 1em plus 0.5em
  minus 0.4em\relax San Francisco, CA, USA: Morgan Kaufmann Publishers Inc.,
  1996, p. 195–210.

\bibitem{STONE1999}
\BIBentryALTinterwordspacing
P.~Stone and M.~Veloso, ``Task decomposition, dynamic role assignment, and
  low-bandwidth communication for real-time strategic teamwork,''
  \emph{Artificial Intelligence}, vol. 110, no.~2, pp. 241 -- 273, 1999.
  [Online]. Available:
  \url{http://www.sciencedirect.com/science/article/pii/S0004370299000259}
\BIBentrySTDinterwordspacing

\bibitem{simmons2007}
\BIBentryALTinterwordspacing
M.~Roth, R.~Simmons, and M.~Veloso, ``Exploiting factored representations for
  decentralized execution in multiagent teams,'' in \emph{Proceedings of the
  6th International Joint Conference on Autonomous Agents and Multiagent
  Systems}, ser. AAMAS '07.\hskip 1em plus 0.5em minus 0.4em\relax New York,
  NY, USA: Association for Computing Machinery, 2007. [Online]. Available:
  \url{https://doi.org/10.1145/1329125.1329213}
\BIBentrySTDinterwordspacing

\bibitem{Oliehoek2013}
F.~A. Oliehoek, S.~Whiteson, and M.~T. Spaan, ``{Approximate solutions for
  factored Dec-POMDPs with many agents},'' \emph{Belgian/Netherlands Artificial
  Intelligence Conference}, pp. 340--341, 2013.

\bibitem{Amato}
C.~{Amato}, G.~{Chowdhary}, A.~{Geramifard}, N.~K. {Üre}, and M.~J.
  {Kochenderfer}, ``Decentralized control of partially observable markov
  decision processes,'' in \emph{52nd IEEE Conference on Decision and Control},
  Dec 2013, pp. 2398--2405.

\bibitem{theile2020uav}
M.~Theile, H.~Bayerlein, R.~Nai, D.~Gesbert, and M.~Caccamo, ``{UAV} path
  planning using global and local map information with deep reinforcement
  learning,'' \emph{arXiv preprint arXiv:2010.06917}, 2020.

\bibitem{bayerlein2021multi}
H.~Bayerlein, M.~Theile, M.~Caccamo, and D.~Gesbert, ``Multi-{UAV} path
  planning for wireless data harvesting with deep reinforcement learning,''
  \emph{IEEE Open Journal of the Communications Society}, vol.~2, pp.
  1171--1187, 2021.

\end{thebibliography}
%



\end{document}